\documentclass[10pt,twocolumn,letterpaper]{article}

\usepackage[pagenumbers]{cvpr} %

\usepackage[dvipsnames]{xcolor}

\def\paperID{1260}
\def\confName{CVPR}
\def\confYear{2025}

\def\modelname{PERSE\xspace}

\def\paperTitle{\textit{\modelname}: Personalized 3D Generative Avatars from A Single Portrait}

\def\authorBlock{
    Hyunsoo Cha \qquad
    Inhee Lee \qquad
    Hanbyul Joo \\
    Seoul National University\\
    {\tt\small \{243stephen, ininin0516, hbjoo\}@snu.ac.kr} \\
    {\tt\small \href{https://hyunsoocha.github.io/perse/}{\color{magenta}{https://hyunsoocha.github.io/perse/}}}
    \vspace{-5mm}
}

\newif\ifreview 
\newif\ifarxiv 
\newif\ifcamera 
\newif\ifrebuttal

\definecolor{tabfirst}{rgb}{1, 0.7, 0.7} %
\definecolor{tabsecond}{rgb}{1, 0.85, 0.7} %
\definecolor{tabthird}{rgb}{1, 1, 0.7} %

\def\champname{\textit{portrait-Champ}\xspace}

\usepackage{times}
\usepackage{microtype}
\usepackage{epsfig}
\usepackage{xcolor}
\usepackage{caption}
\usepackage{float}
\usepackage{placeins}
\usepackage{color, colortbl}
\usepackage{stfloats}
\usepackage{enumitem}
\usepackage{tabularx}
\usepackage{xstring}
\usepackage{multirow}
\usepackage{xspace}
\usepackage{cuted} %
\usepackage{url}
\usepackage{subcaption}
\usepackage{xcolor}
\usepackage[hang,flushmargin]{footmisc}
\usepackage{bm}
\usepackage{amsmath}
\usepackage{booktabs} %
\usepackage{makecell} %
\usepackage{amssymb}
\ifcamera \usepackage[accsupp]{axessibility} \fi
\ifcamera  \fi
\newcommand{\ourmethod}[1]{OFHR}

\ifarxiv  \fi

\renewcommand{\vec}[1]{\bm{#1}}

\newcommand{\Mat}[1]{\mathbf{#1}}
\newcommand{\R}[1]{{%
    \textbf{%
        \ifstrequal{#1}{1}{\textcolor{orange}{R#1}}{%
        \ifstrequal{#1}{2}{\textcolor{blue}{R#1}}{%
        \ifstrequal{#1}{3}{\textcolor{magenta}{R#1}}{%
        \ifstrequal{#1}{4}{\textcolor{teal}{R#1}}{%
                           \textcolor{cyan}{R#1}%
        }}}}%
    }%
}}

\newcommand{\figref}[1]{Fig.~\ref{#1}}
\newcommand{\tabref}[1]{Tab.~\ref{#1}}
\newcommand{\eqnref}[1]{Eq.~(\ref{#1})}
\newcommand{\secref}[1]{Sec.~\ref{#1}}

\usepackage{subcaption} %
\definecolor{deepblue}{RGB}{83, 124, 186}

\definecolor{cvprblue}{rgb}{0.21,0.49,0.74}
\usepackage[pagebackref,breaklinks,colorlinks,citecolor=cvprblue]{hyperref}
\usepackage{cuted}

\title{\paperTitle}

\author{\authorBlock}
\begin{document}
\maketitle
\begin{strip}\centering
\includegraphics[width=\linewidth, trim={0 0cm 0 0.0cm}, clip]{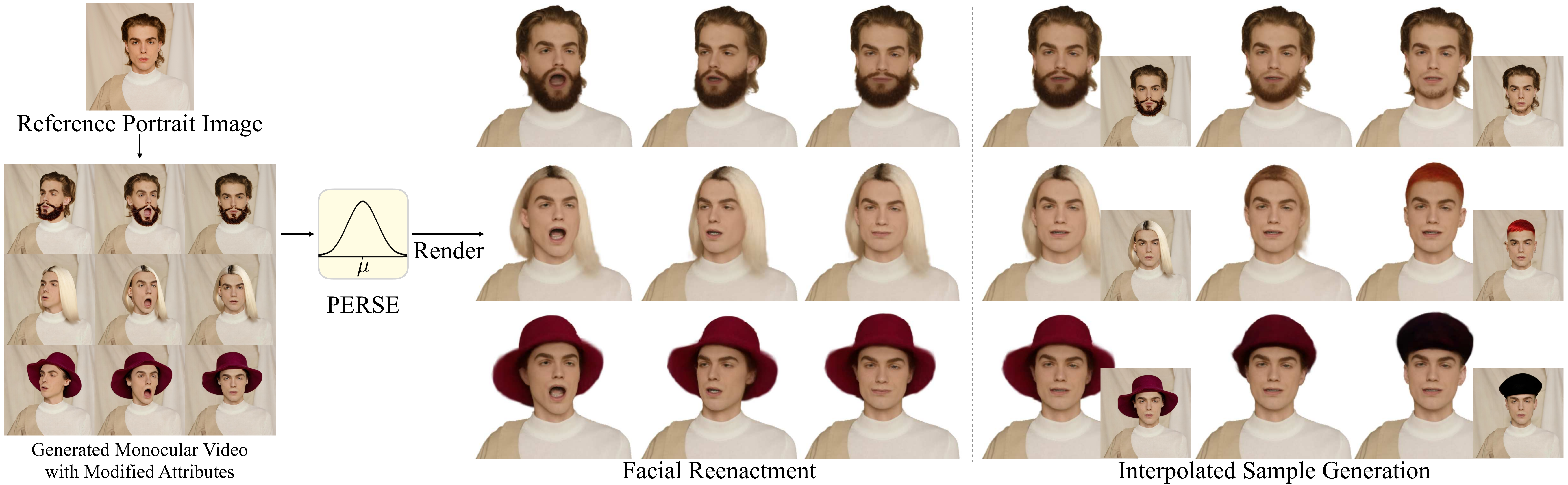}
\vspace{-13px}
\captionof{figure}{\textbf{\modelname.} Given a reference portrait image as input, our method constructs an animatable  personalized 3D generative avatar with disentangled and editable control over various facial attributes such as beard, hair, and hat.
}
\label{fig:teaser}

\end{strip}
\begin{abstract}
We present \modelname, a method for building a personalized 3D generative avatar from a reference portrait. Our avatar enables facial attribute editing in a continuous and disentangled latent space to control each facial attribute, while preserving the individual's identity. To achieve this, our method begins by synthesizing large-scale synthetic 2D video datasets, where each video contains consistent changes in facial expression and viewpoint, along with variations in a specific facial attribute from the original input. We propose a novel pipeline to produce high-quality, photorealistic 2D videos with facial attribute editing. Leveraging this synthetic attribute dataset, we present a personalized avatar creation method based on 3D Gaussian Splatting, learning a continuous and disentangled latent space for intuitive facial attribute manipulation. To enforce smooth transitions in this latent space, we introduce a latent space regularization technique by using interpolated 2D faces as supervision. Compared to previous approaches, we demonstrate that \modelname generates high-quality avatars with interpolated attributes while preserving the identity of the reference individual.
\end{abstract}
    
\section{Introduction}
\vspace{-5px}
A personalized 3D face avatar can represent each individual in VR/AR environments, replicating the user's appearance and facial expressions. However, the \emph{exact replication} of the appearance does not fully reflect real-world humans. In reality, people often change the attributes of their appearance, like hairstyles, or start growing a beard or mustache. Users may also wish to adjust their facial features in the virtual world, like the shape of their nose, eyebrows, or mouth, enhancing their desired look while preserving their core identity. While most prior avatar creation methods focus on building an exact digital twin of the person from images or video data~\cite{zheng2022avatar, zheng2023pointavatar, chen2024monogaussianavatar, zielonka2023instant, gafni2021dynamic, saito2024relightable, shao2024splattingavatar, xu2023latentavatar}, the personalized avatar model with the generative ability to control and edit facial attributes remains underexplored. 

In this work, we present \modelname, a method to build an animatable personalized 3D generative avatar from a single reference portrait image. 
Our method goes beyond merely creating an exact twin from video inputs, introducing a novel approach that emphasizes flexibility and control over facial attributes, such as changing hairstyles or beards shown in \figref{fig:teaser}. 
To build \modelname from a single reference portrait image, we generate a large-scale 2D monocular synthetic video dataset of the reference identity, where each video has a variation in a specific facial attribute from the original input (e.g., a different hairstyle) driven by the face motion guidance, as shown in \figref{fig:synthetic_dataset}. Each video is also paired with a text prompt description addressing the changed attribute. To build this high-quality, photorealistic synthetic video dataset, we introduce a new pipeline that begins with synthesizing 2D images with attribute editing in a fully automated procedure. This is followed by a portrait animation process that leverages a combination of an existing pre-trained 2D portrait animation method~\cite{guo2024liveportrait} and our newly trained image-to-video model extending a prior work~\cite{zhu2024champ}. 
Notably, our synthetic video generation process is efficient, scalable, and provides significantly more attribute diversity by effectively synthesizing a thousand attribute videos compared to tens in prior work~\cite{cha2024pegasus}. Using this synthetic video dataset, we train an avatar model with the continuous and disentangled attribute latent space.

To enhance the generative ability of our avatar model for unseen or interpolated attribute appearances, we also present a novel technique to enforce continuous and smooth latent space. To achieve this, we present a latent space regularization technique by using interpolated 2D face images from an image morphing technique~\cite{zhang2024diffmorpher} (e.g., synthesizing medium-length hair from short hair and long hair attributes), providing pseudo supervisions for the interpolated latent spaces.
We show the efficacy of our regularization technique by producing novel and unseen attributes from interpolated latent spaces, as shown in \figref{fig:qual_baselines}. Furthermore, we present an efficient fine-tuning technique via Low-Rank Adaptation (LoRA)~\cite{hu2021lora}, to integrate any new facial attributes from in-the-wild images into our avatar model. 

Our contributions are summarized as follows: 
(1) the first method to generate an animatable 3D personalized generative avatar from a reference portrait image with controllable facial attributes;
(2) a method to generate high-quality synthetic 2D video datasets with diverse attribute editing from a reference portrait image;
(3) latent space regularization by using face morphing supervision for continuous and smooth latent space to enhance the generative ability for unseen or interpolated attribute appearances;
(4) an efficient fine-tuning technique via Low-Rank Adaptation (LoRA)~\cite{hu2021lora} to integrate any new facial attribute into the avatar model.
\vspace{-5px}

\section{Related Work}
\vspace{-5px}
\noindent\textbf{3D Facial Avatar Reconstruction.}
Since the introduction of foundational 3D Morphable Model~\cite{blanz19993dmm} (3DMM), parametric 3D face models~\cite{blanz19993dmm, li2017learning, cao2013facewarehouse} have evolved to capture the diverse and dynamic nature of human faces, representing variations in shape, head pose, and facial expression through a set of parameters. 
Building on these models, various methods reconstruct 3D face avatars from single portrait images by estimating 3DMM parameters~\cite{sanyal2019learning, feng2021learning, danvevcek2022emoca}. 
Recently, monocular 3D avatar reconstruction methods~\cite{gafni2021dynamic, zheng2022avatar, zheng2023pointavatar, chen2024monogaussianavatar, kirschstein2024diffusionavatars, lee2024guess} generate morphable photorealistic avatars leveraging advancements in 3D representation~\cite{mildenhall2021nerf,kerbl20233d}. 

To move beyond single-subject avatars, the PEGASUS~\cite{cha2024pegasus} reconstructs a personalized 3D generative avatar enabling control over facial attributes while preserving the reference identity, using synthetic DB. Similarly, HeadGAP~\cite{zheng2024headgap} trains a generalizable prior model for 3D head avatar leveraging a large-scale multiview dataset and an avatar model with part-specific and point-specific feature codes. 
Despite advancements, constructing a unified 3D representation that can precisely capture and control all facial attributes remains challenging.
To address this, disentangled or hybrid representations have been proposed, enabling selective modification of facial features or garments~\cite{feng2023learning, feng2022capturing, kim2024gala}. 
However, these approaches are limited by discrete 3D structures, restricting continuous interpolation capabilities.
Recently, latent-conditioned generative models~\cite{kim2023ncho, li2023megane, ho2023learning, cha2024pegasus} have been introduced to mitigate these constraints, yet they often lack the capacity for fine-grained editing and are confined to specific categories.

\noindent\textbf{Smooth Image Morphing and Interpolation.}
Generating a plausible intermediate image between two pivot images has been widely studied within the context of image generative field~\cite{Karras2021sgan3, song2020denoising, ho2020denoising}.
The recent breakthroughs of diffusion model~\cite{song2019generative, ho2020denoising, rombach2022high} improved the image interpolation methods to generate more plausible and better quality interpolated images with less limitation on categories~\cite{wang2023interpolating, zhang2024diffmorpher, Guo_2024_smooth, zhang2024tvg, shen2024dreammover, song2019generative, zheng2024noisediffusion}.
Many diffusion-based interpolation methods follow the procedure of DDIM inversion~\cite{song2020denoising, mokady2023null}, interpolation in diffusion latent space, and DDIM forward sampling with slight modification.
DiffMorpher~\cite{zhang2024diffmorpher} additionally utilizes personalized diffusion models finetuned on each pivot image with LoRA~\cite{hu2021lora} to produce smooth interpolated sequences.
SmoothDiffusion~\cite{Guo_2024_smooth} finetunes diffusion models with LoRA~\cite{hu2021lora} to preserve the distance of interpolated sample and pivots during denoising.

\noindent\textbf{Portrait Animation from Single Image.}
Generating animations from a single image is a challenging task that has seen significant advancements through generative models, particularly based on implicit keypoints and diffusion methods.

Several approaches~\cite{siarohin2019first, wang2021one, siarohin2021motion, mallya2022implicit, zhao2022thin, hong2022depth} have introduced intermediate motion representations based on implicit keypoints estimation, enabling the mapping of a source portrait image to a driving image using optical flow. 
Extending the previous work~\cite{wang2021one}, LivePortrait~\cite{guo2024liveportrait} enhances animation quality by integrating a GAN-based decoder~\cite{park2019semantic}, resulting in effective and controllable portrait animations.

Recent advancements in diffusion models have significantly enhanced portrait animation, offering improved control and realism. Several methods~\cite{xu2024magicanimate, hu2024animate, chang2023magicpose} have explored full-body animations guided by motion sequence drived from body keypoints. Building upon previous approaches~\cite{hu2024animate}, Champ~\cite{zhu2024champ} generates full-body animations guided by multiple reference videos such as SMPL~\cite{SMPL:2015} renderings. 

\section{Preliminary: PEGASUS~\cite{cha2024pegasus}}
\vspace{-5px}
\begin{figure*}[t]\centering
\includegraphics[width=\linewidth, trim={0 0 0 0},clip]{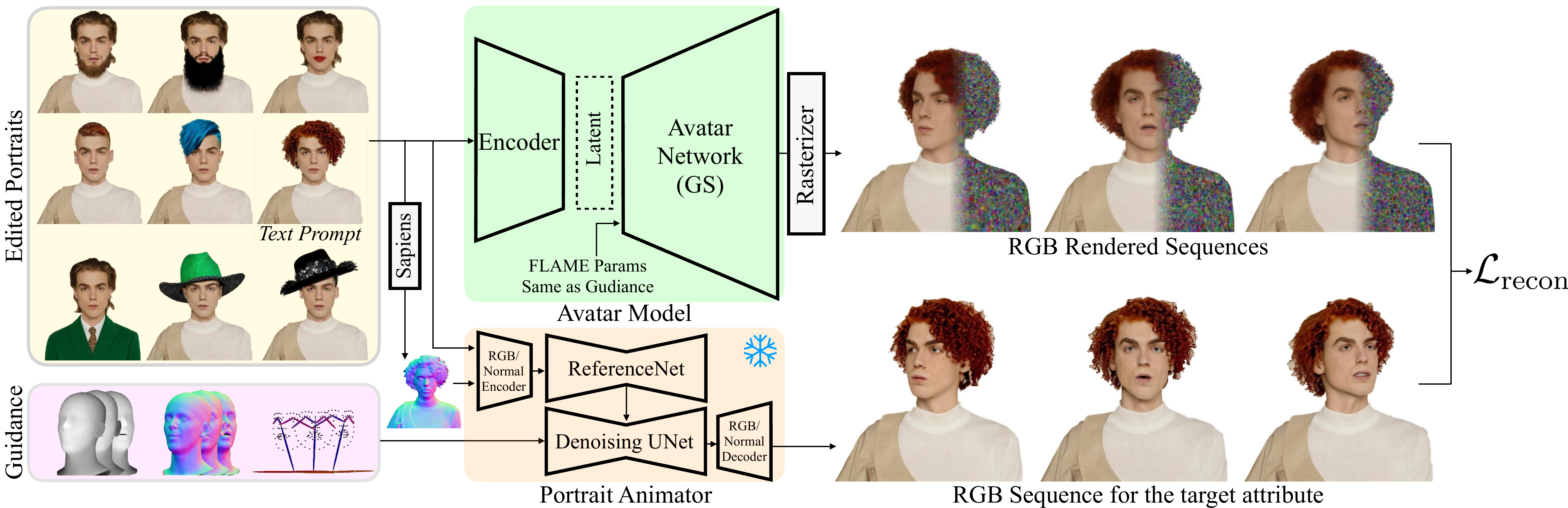}
\vspace{-15px}
\captionof{figure}{\textbf{Overview of Synthetic Dataset Generation and Avatar Model Training.} 
Starting with a collection of edited portrait images, we generate RGB videos for each target attribute using Portrait Animator. 
The guidance for the Portrait Animator is derived from tracked FLAME parameters of a predefined training motion sequence, which also serve as inputs to the avatar network in our avatar model. 
Using the generated RGB videos, we train our avatar model with a reconstruction loss. Each edited portrait is paired with the text prompt used for its generation. The process of creating these edited portraits based on text prompts is detailed in \secref{sec:synthetic_dataset_generation}. 
}
\label{fig:method_stage1}
\vspace{-10px}
\end{figure*}

Our avatar model is based on a previously proposed personalized generative 3D avatar, PEGASUS~\cite{cha2024pegasus}
, by modifying its original 3D point cloud representation~\cite{zheng2023pointavatar} into 3D Gaussian Splatting~\cite{kerbl20233d}.
The PEGASUS avatar model is an animatable 3D avatar model of a reference individual with disentangled controls to selectively alter facial attributes such as hair or nose, while preserving the reference identity. 
The PEGASUS model takes a latent code $\mathbf{z} \in \mathbb{R}^{(N_c+1) \times d}$ along with FLAME parameters $\beta$ (shape), $\theta$ (pose), and $\psi$ (expression) as inputs, and infers a colorized point cloud to express the target individual's appearance, pose and expressions changes: 
\vspace{-3px}
\begin{equation}\label{eq:network_io}
    \{ \mathbf{x}_i^{d}, \mathbf{n}^d, \mathbf{a}_i \} = \mathcal{M}_{\phi}(\mathbf{x}_i^{gc}, \mathbf{z}, \vec{\beta}, \vec{\theta}, \vec{\psi}),
\end{equation}
where $\mathbf{x}_i^{d}$ is the 3D point locations, $\mathbf{n}_i\in \mathbb{R}^3$ is the point normals, and  $\mathbf{a}_i \in \mathbb{R}^3$ is the point albedo colors.
The input latent code $\mathbf{z} \in \mathbb{R}^{(N_c+1) \times d}$ is a concatenation of $N_c$ subpart latent codes $\{\mathbf{z}_j\}_{j=0...N_c}$, where each subpart latent code $\mathbf{z}_j \in \mathbb{R}^d$ controls specific aspects of the human identity or a subpart. 
The identity latent code $\mathbf{z}_0$ controls overall identity variations, and the other latent codes $\mathbf{z}_{j\neq0}$ control each subpart, preserving the identity defined by $\mathbf{z}_0$. 

Notably, the PEGASUS model relies on constructing a synthetic video collection of the reference identity with edited facial attributes. 
This is performed by replacing specific facial attributes in the reference person's video with those from multiple other individuals' videos. 
Consequently, building the synthetic dataset requires not only the video of the reference individual but also numerous videos from other individuals for attribute variations. 
Moreover, this approach involves a time-intensive process of creating 3D avatars for each individual to synthesize all attribute variations, which limits the scalability of the method.

\section{Method}
\vspace{-3px}
We first describe our personalized 3D generative avatar model creation (\secref{sec:personalized_generative_avatar_model}), extending the previous work~\citep{cha2024pegasus}. Then, we introduce our pipeline for generating a large-scale synthetic 2D facial attribute dataset  (\secref{sec:synthetic_dataset_generation}). Additionally, we present our novel training scheme including latent space regularization with interpolated 2D faces (\secref{sec:method_training}), and also present our efficient fine-tuning technique to integrate arbitrary new attributes into our optimized latent space while preserving the existing distribution (\secref{sec:method_lora}).

\subsection{Personalized Generative Avatar Model}\label{sec:personalized_generative_avatar_model}
\vspace{-3px}
\noindent\textbf{3D Gaussian Splatting for Avatar.}
Our avatar model builds on the structure of PEGASUS~\cite{cha2024pegasus} with several modifications.
First, we change the 3D representation of the avatar from a colorized point cloud to 3D Gaussian Splatting (3D-GS)~\cite{kerbl20233d} which enhances rendering quality. This is achieved by estimating 3D Gaussian parameters for each point, replacing the original point normal and albedo.
Specifically, our model takes a latent code $\mathbf{z}$ and FLAME parameters $\{\vec{\beta}, \vec{\theta}, \vec{\psi}\}$ as inputs, and 
infers 3D Gaussian parameters of posed avatar, including the 3D position $\mathbf{x}_i^{d} \in \mathbb{R}^3$, rotation $\mathbf{r}^{d}_i \in \mathbb{R}^4$, scale $\mathbf{s}^{d}_i \in \mathbb{R}^3$, opacity $\mathbf{o}^{d}_i \in \mathbb{R}$, and color $\mathbf{c}_i \in \mathbb{R}^3$ as follows:
\begin{equation}\label{eq:avatar_model}
    \{\vec{x}_i^{d}, \vec{r}_i^{d}, \vec{s}_i^{d}, \vec{o}_i^{d}, \vec{c}_i^{d} \} = \mathcal{M}_{\Theta}(\mathbf{x}_i^{gc}, \mathbf{z}, \vec{\beta}, \vec{\theta}, \vec{\psi}).
\end{equation}
To capture fine-grained deformations conditioned on head pose, we introduce an additional MLP deforming 3D Gaussians based on the input FLAME parameters $\{\vec{\beta}, \vec{\theta}, \vec{\psi}\}$, similar to MonoGaussianAvatar~\cite{chen2024monogaussianavatar}. 
We densify the 3D Gaussians to capture fine detail using the upsampling strategy of prior work~\cite{cha2024pegasus, zheng2023pointavatar} and prune distracting Gaussians through opacity resetting and thresholding as in the original 3D-GS framework~\cite{kerbl20233d}.
By rasterizing the 3D Gaussians, we get a rendering of the avatar as follows:
\vspace{-1px}
\begin{equation}\label{eq:rasterize}
    \hat{\Mat{I}} = \text{GSR} \Big(\{\vec{x}_i^d, \mathbf{r}_i^d, \vec{s}_i^{sc}, \vec{o}_i^{d}, \vec{c}_i^{sc}\}_{i \in \{ 1 \cdots N \} } \Big),
\end{equation}
where GSR represents a 3D-GS Rasterizer~\cite{kerbl20233d}.

\noindent\textbf{CLIP-guided Latent Space Configuration.} 
Following PEGASUS~\cite{cha2024pegasus} model, we represent our avatar model latent code $\mathbf{z} \in \mathbb{R}^{N_c \times d}$ as a concatenation of $N_c$ subpart latent codes $\{\mathbf{z}_j\in\mathbb{R}^d\}_{j=[1\cdots N_c]}$. 
This part-wise separated latent configuration allows to control each facial attribute while preserving other facial attributes. 
We can also selectively transfer the target attribute of the $k$-th subpart, such as hair, to the reference avatar by substituting the $k$-th subpart latent as follows:
\begin{equation}
    \mathbf{z}^{\text{new}}_{j} = \begin{cases}
    \mathbf{z}^{\text{ref}}_{j} & \text{if } j \neq k \\
    \mathbf{z}^{\text{target}}_{j} & \text{if } j = k
\end{cases}
\end{equation}
To achieve this disentangled latent space, we optimize a single reference latent code 
$\mathbf{z}^\text{ref} \in \mathbb{R}^{N_c \times d}$, representing the identity of the input portrait image, along with a set of subpart latent codes $\{\mathbf{z}_{k}^\text{target}\in\mathbb{R}^d\}$, where each corresponds to a specific subject in our synthetic dataset.  

However, directly optimizing these latent codes $\{\mathbf{z}_{k}^\text{target}\in\mathbb{R}^d\}$ are prone to be overfitted on each subject, resulting in poor generalization to unseen subjects.
To address this and achieve more compact latent space, we constrain latent codes using a well-established text-image feature model CLIP~\cite{radford2021learning}, which is a key difference over previous work~\cite{cha2024pegasus}. 
We define the target subpart latent as an output of shallow MLP network conditioned on CLIP image and text features $f_{I},f_{T} \in \mathbb{R}^{512}$:
\begin{equation}\label{eq:latent_mapping_mlp}
    \mathbf{z}_k^\text{subject} = \text{MLP}_\text{z}(f_{I}, f_{T}).
\end{equation}
The CLIP features are calculated from front-view reference synthetic image and text pairs from our synthetic datasets. Additionally, we define $z_\text{zero}$ as a unique shared subpart latent code representing an empty subpart, such as the absence of a hat or beard.

\subsection{Synthetic Dataset Generation}\label{sec:synthetic_dataset_generation} 
\vspace{-5px}
We create a synthetic portrait video dataset with varying facial attributes from the input image of the reference individual to enable the generative ability for our 3D avatar model. Our synthetic dataset generation pipeline is performed via a two-stage process: generating attribute-edited images and animating the edited portrait images.

\noindent\textbf{Attribute-Edited Portrait Image Generation.}
Given a portrait image $\Mat{I}_\text{input}$ of the reference individual, our goal is to photo-realistically edit each attribute to reflect a different style. We consider 9 attribute categories: beard, clothes, earrings, eyebrows, hair, hat, headphones, mouth, and nose.
To achieve this goal, we present a text-conditioned image inpainting pipeline by leveraging multiple tools including pre-trained 2D diffusion models~\cite{flux_controlnet_alpha}.
We first determine a list of text prompts for each attribute category with specific adjectives (e.g., curly, straight, and wavy for hair). We leverage ChatGPT~\cite{openai2024chatgpt} to explore various possible distinctive adjectives.
Then, for each text description $T$, we synthesize a corresponding portrait image with attribute changes using a text-conditioned inpainting model~\cite{flux_controlnet_alpha}: 
\begin{equation}
    \Mat{I}_\text{gen} = \text{I2I}_\text{inpaint}(\Mat{I}_\text{input}, T, \Mat{M}_\text{edit}),
    \label{eq:inpaint}
\end{equation}
where $\Mat{M}_\text{edit}$ denotes the mask region where the inpainting module needs to modify. 
Importantly, we find providing a suitable mask region $\Mat{M}_\text{edit}$ is essential to synthesizing photo-realistic output that adheres to the text guidance. A segmentation mask directly derived from the original input typically results in minor color changes without substantial shape variations.

\begin{figure}[t]
\includegraphics[trim={0 0 0 0},clip,width=\columnwidth]{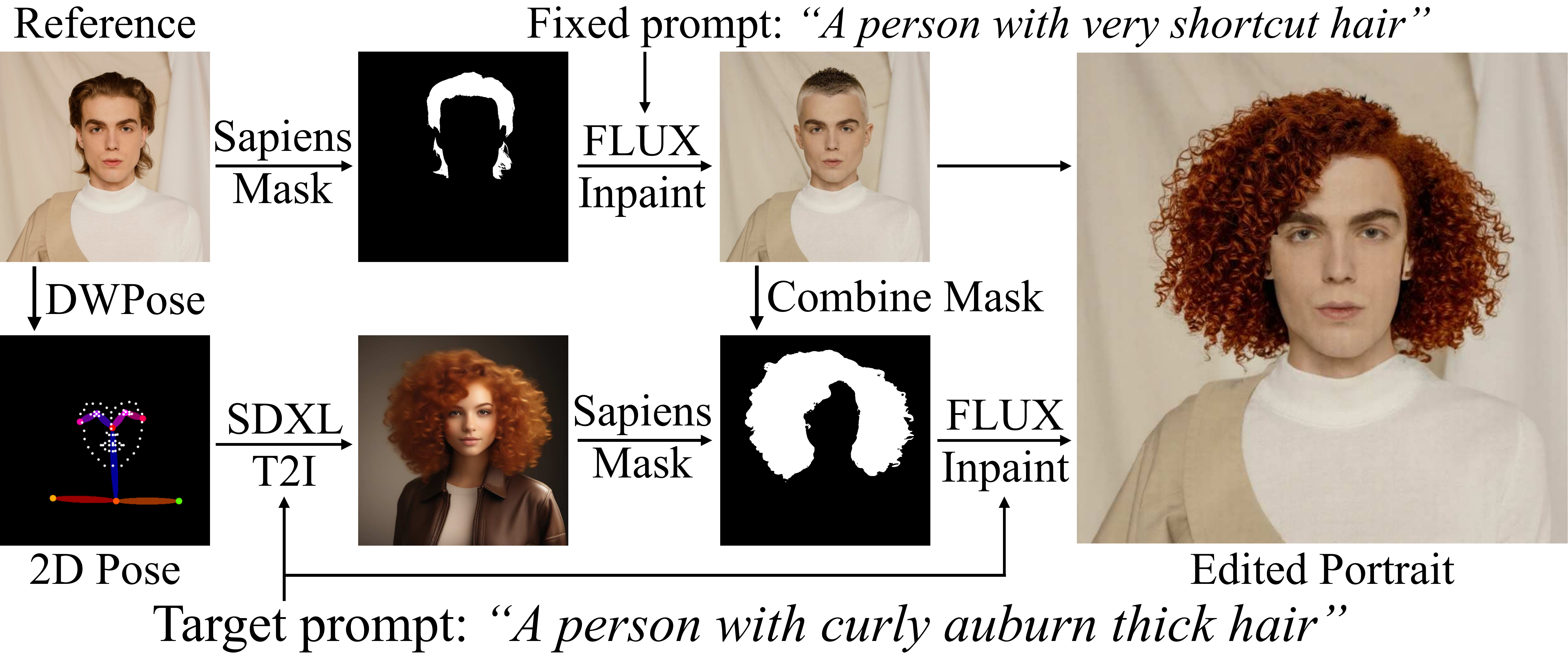}
\vspace{-18px}
\caption{\textbf{Image Synthesis.} 
Starting from a reference portrait image, we present a fully automatic pipeline that generates an edited portrait without any manual manipulation such as user scribbles. To automatically generate the optimal mask image for inpainting, our method leverages SDXL, Sapiens, and FLUX~\cite{flux_controlnet_alpha, podell2023sdxl, khirodkar2025sapiens}.
}
\label{fig:image_synthesis}
\vspace{-15px}
\end{figure} 

\begin{figure}[t]
\includegraphics[trim={0 0 0 0},clip,width=\columnwidth]{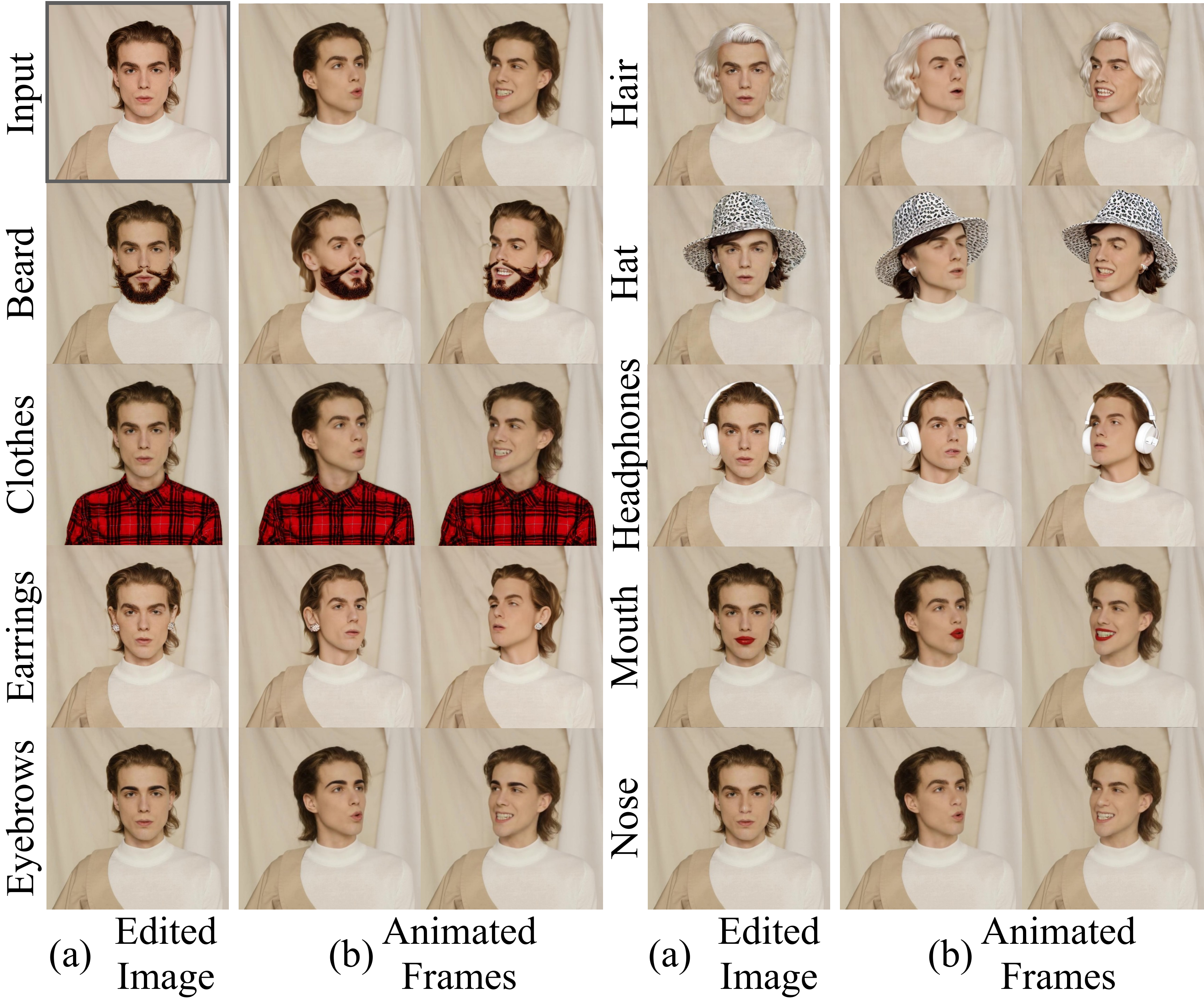}
\vspace{-20px}
\caption{\textbf{Our Synthetic Dataset.} 
The upper left black bounded image is the input portrait. (a) is an edited image from the input portrait and (b) is a generated frame by the portrait animator.
}
\label{fig:synthetic_dataset}
\vspace{-15px}
\end{figure}

To generate mask images $\Mat{M}_\text{edit}$ that are optimally aligned with a given text prompt $T$, we introduce a fully automatic image synthesis pipeline.
Specifically, we synthesize a new portrait image $\Mat{I}_\text{text}$ from the text $T$ using a text-to-image diffusion model~\cite{podell2023sdxl}, where we enforce the facial poses and expressions of the synthesized image align with the $\Mat{I}_\text{input}$ using ControlNet~\cite{zhang2023adding}:  
\begin{equation}
    \Mat{I}_\text{text} = \text{T2I} \big(T, C(\Mat{I}_\text{input})\big),
\end{equation} 
where $C(\Mat{I}_\text{input})$ is the OpenPose~\cite{cao2017realtime} keypoint image obtained by applying off-the-shelf keypoint estimator~\cite{yang2023effective} on $\Mat{I}_\text{input}$. 
Although the identity of $\Mat{I}_\text{text}$ is not necessarily the same as $\Mat{I}_\text{input}$, its facial pose is aligned to the $\Mat{I}_\text{input}$, allowing us to obtain the attribute mask $\Mat{M}_\text{edit}$ accordingly. 
We extract the attribute mask $\Mat{M}_\text{text}$ from $\Mat{I}_\text{text}$ using an off-the-shelf segmentation network~\cite{khirodkar2025sapiens} and use it as the target area to edit $\Mat{M}_\text{edit} = \Mat{M}_\text{text}$ for \eqnref{eq:inpaint}. Examples are shown in the first column of \figref{fig:synthetic_dataset}.

For attributes of hat and hair, an additional step is required to remove the original parts that may unexpectedly remain after the inpainting process (e.g., the case that the original hair shape is bigger than $\Mat{M}_\text{text}$). 
We resolve this issue by editing the original input image $\Mat{M}_\text{input}$ with a version containing shortcut hair, denoted  $\Mat{M}_\text{shortcut}$, before applying inpainting: 
\begin{equation}
    \Mat{I}_\text{shortcut} = \text{I2I}_\text{inpaint}( \Mat{I}_\text{input}, T_\text{shortcut}, \Mat{M}_\text{input}),
\end{equation}
where $\Mat{M}_\text{input}$ is the hair mask of the $\Mat{I}_\text{input}$, and $T_\text{shortcut}$ is a corresponding text prompt: ``A person with very shortcut hair''.
We also find that, in these categories, combining the mask of this shortcut hair image $\Mat{I}_\text{shortcut}$ with the mask from the text-to-image output $\Mat{M}_\text{text}$ produces superior results with fewer artifacts:
\begin{equation}
    \Mat{M}_\text{edit} = (\Mat{M}_\text{shortcut} \cup \Mat{M}_\text{text}).
\end{equation}
See Fig.~\ref{fig:image_synthesis} for the overview of the editing pipeline.

\begin{figure}[t]
\includegraphics[trim={0 0 0 0},clip,width=\columnwidth]{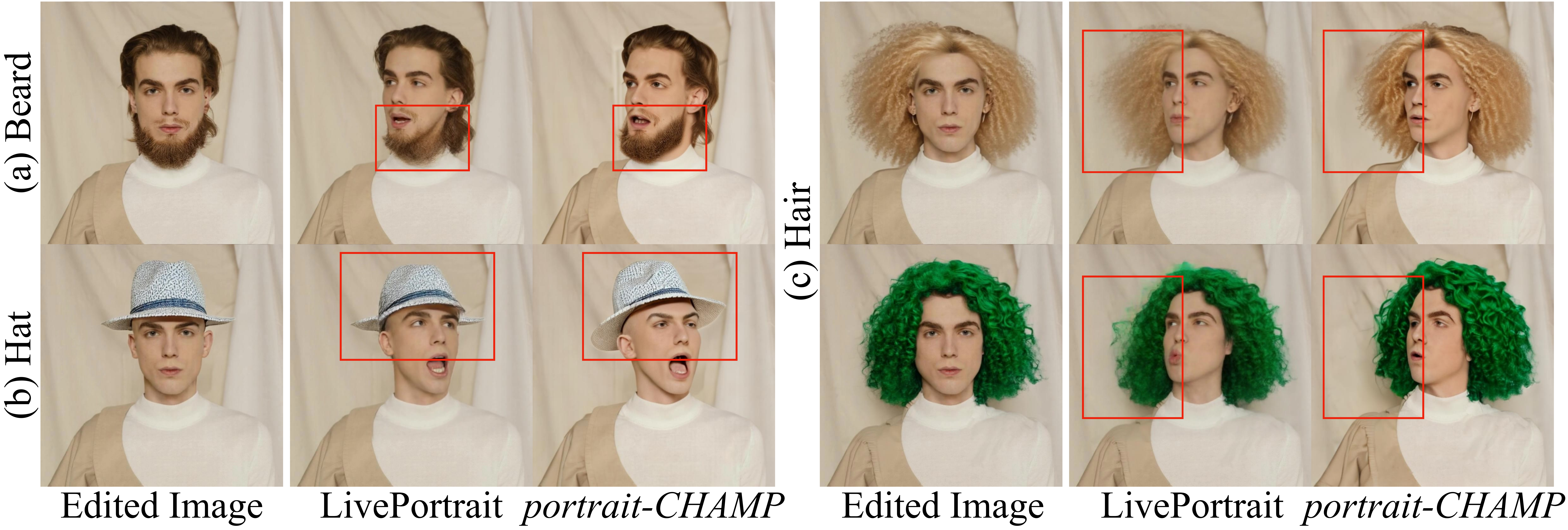}
\vspace{-18px}
\caption{\textbf{Comparison of LivePortrait and \champname.} Examples from LivePortrait~\cite{guo2024liveportrait} and \champname~demonstrate several limitations: (a) artifacts are visible in the hair region, (b) LivePortrait lacks adaptability to head poses involving hats, and (c) beard artifacts are prone to aliasing and disappearance.}
\label{fig:example_liveportrait_portraitchamp}
\vspace{-15px}
\end{figure}    

\noindent\textbf{Animated Portrait Video Generation.}
We animate each edited portrait image $\Mat{I}_\text{gen}$ to synthesize a video with varying head poses and facial expressions, which are used as a pseudo monocular video dataset for training our animatable 3D personalized generative avatar model. 

To achieve this goal, we utilize two different portrait animation techniques, LivePortrait~\cite{guo2024liveportrait} and our customized face-specialized Champ~\cite{zhu2024champ}: \champname. These methods are chosen for their complementary strengths. 
The goal of both animators is the same producing a video output following the motion guidance while preserving the identity given by the input image:
\begin{equation}
    \Mat{V}_\text{gen} = \text{I2V}(\Mat{I}_\text{gen}, \mathcal{G}),
    \label{eq:portrait_anim}
\end{equation}
where $\mathcal{G}$ denotes a set of motion guidance, including the FLAME depth map, FLAME normal map, and 2D body and facial keypoints, as shown in the guidance at \figref{fig:method_stage1}.
To obtain $\mathcal{G}$, we capture a short video with varied head poses and facial expressions, and apply a monocular face capture method~\cite{danvevcek2022emoca} to extract FLAME parameters, from which we extract the motion guidance cues $\mathcal{G}$. The same  $\mathcal{G}$ is used for all generated videos, $\Mat{V}_\text{gen}$, resulting in a collection of videos with the same motions and diverse attribute changes. Examples are shown in \figref{fig:synthetic_dataset}.

For attribute categories excluding beard, earrings, hair, hat, and headphones, we use LivePortrait~\cite{guo2024liveportrait} to animate the edited portrait images.
Although LivePortrait successfully generates high-quality face-animation videos, it performs suboptimally with certain attributes and conditions. For example, with portrait images featuring voluminous hair, long beards, or large hats, particularly in cases with extensive head movements, LivePortrait model often generates unnatural deformations, such as stretching and shrinking with noticeable artifacts as shown in \figref{fig:example_liveportrait_portraitchamp}.

To address these limitations, we build and train our own alternative image-to-video diffusion model, \champname to leverage high temporal consistency of 2D video diffusion model~\cite{zhu2024champ, guo2023animatediff}.
Our model shows superior performance for synthesizing beard, earrings, hair, hat, and headphones over LivePortrait~\cite{guo2024liveportrait}, as demonstrated in our experiments. 
We build our model based on the Champ~\cite{zhu2024champ} that is originally designed for full-body animations, with a few extensions.
For concise control of head and facial expression, our \champname inputs normal and depth rendering of EMOCA~\cite{danvevcek2022emoca} as conditioning input.
We add a normal channel in VAE encoder and decoder of \champname to enhance 3D-awareness of the video diffusion model~\cite{he2024magicman}, and trained it with 6k real-world videos capturing diverse identities and motions~\cite{yu2023celebv}.

\begin{figure*}[t]\centering
\includegraphics[width=\linewidth, trim={0 0 0 0},clip]{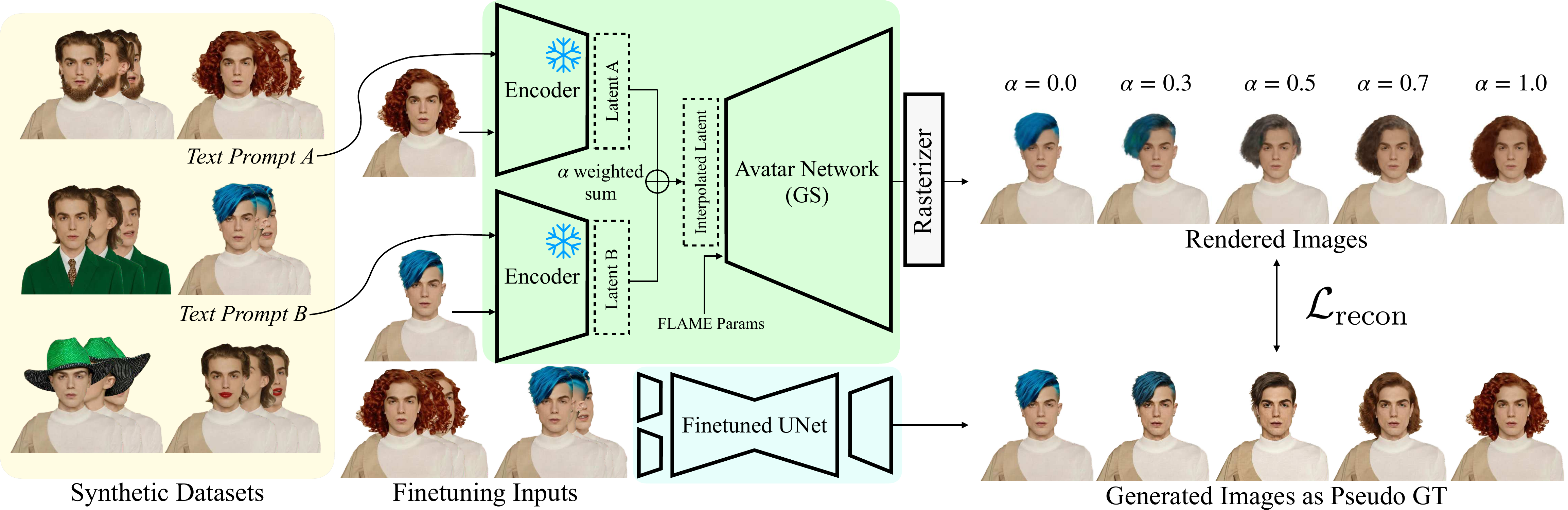}
\vspace{-10px}
\captionof{figure}{\textbf{Overview of Supervision for Interpolation.} We propose an additional training strategy that leverages a finetuned 2D diffusion model~\cite{zhang2024diffmorpher, hu2021lora} to enhance 
the quality of interpolated samples in latent space. 
Starting from two samples with text prompts A and B, 
we generate interpolated latent codes through weighted summation based on $\alpha$. We then compute the part-wise loss and backpropagate it through the avatar model.}
\label{fig:method_stage2}
\vspace{-15px}
\end{figure*}

\subsection{Training}\label{sec:method_training}
\vspace{-5px}
In essence, training our avatar model on the synthetic dataset is identical to the process of reconstructing a 3D avatar from real 2D video inputs. At each iteration, we render an image $\hat{\Mat{I}}$ of a posed subject from the synthetic dataset and calculate the reconstruction loss, $\mathcal{L}_\text{recon}$, by comparing it to the ground truth image, $\Mat{I}$.
\begin{equation}\label{eq:recon_loss}
    \begin{aligned}
        \mathcal{L}_{\text{recon}}(\hat{\Mat{I}}, \Mat{I}) = 
        &\ \lambda_{\text{L1}}||\hat{\Mat{I}}-\Mat{I}||_1 \\
        &+ \lambda_{\text{SSIM}}\text{SSIM}(\hat{\Mat{I}}, \Mat{I}) 
        + \lambda_{\text{VGG}}\text{VGG}(\hat{\Mat{I}}, \Mat{I})
    \end{aligned}
\end{equation}
We then compute latent regularization loss $\mathcal{L}_\text{z}$ enforcing the norm of the latent code close to be zero and estimated FLAME parameters regularizing loss $\mathcal{L}_\text{FLAME}$ following PEGASUS~\cite{cha2024pegasus}. 
Our total loss is as follows:
\begin{equation}\label{eq:total_loss}
    \mathcal{L}_{\text{tot}} = 
    \lambda_{\text{recon}}\mathcal{L}_{\text{recon}}\big(\hat{\Mat{I}}, \Mat{I}\big)  +\lambda_{\text{FLAME}}\mathcal{L}_{\text{FLAME}} 
    +\lambda_{\text{z}}\mathcal{L}_{\text{z}}.
\end{equation}
We train our model with this objective until convergence.
\noindent\textbf{Fine-tuning for Interpolated Samples.}
After convergence, our avatar model still suffers from sampling high quality avatar which is not included in the trained subject. 
The sampled avatars frequently contain artifacts, such as floating Gaussians or unnatural color blobs as illustrated in \figref{fig:results_interpolation}.
To mitigate these artifacts, we propose an interpolation regularization loss leveraging prior knowledge from a pretrained image diffusion model~\cite{rombach2022high}, as demonstrated in \figref{fig:method_stage2}. 
By regularizing the interpolated renderings to be closer to image generated by the diffusion interpolation generator~\cite{zhang2024diffmorpher}, we improve both the rendering quality and realism of interpolated samples. 

To calculate the interpolation loss, we sample two pivot subjects, $(a,b)$ from the same category in our synthetic dataset and render an interpolated subject in every iteration:
\begin{equation}\label{eq:interp_sample}
    \hat{\Mat{I}}_{\text{interp}, \alpha} = \text{GSR}\Big(\mathcal{M}_\Theta(\mathbf{z}^{a}(1-\alpha) + \mathbf{z}^{b}\alpha)\Big),
\end{equation}
where $\alpha$ denotes an interpolation weight. 
We use DiffMorpher~\cite{zhang2024diffmorpher} to generate semantically plausible and visually realistic interpolations between their images, controlled by the same interpolation weight $\alpha$:
\begin{align}\label{eq:interp_Diffmopher}
    \Mat{I}_{\text{interp}, \alpha} &= \text{DiffMorpher}_{\alpha}\Big(\Mat{I}_{a}, \Mat{I}_{b}\Big).
\end{align}
As DiffMorpher~\cite{zhang2024diffmorpher} generated image $\Mat{I}_{\text{interp}, \alpha}$ often fails to preserve identity, we apply loss only on the subpart region $M_{\text{part}}$ which alters during interpolation: 
\begin{equation} \label{eq:interpolation_loss}
    \mathcal{L}_\text{interp} = \mathcal{L}_{\text{recon}}\Big(\Mat{M}_{\text{part}}\circ \Mat{I}_{\text{interp}, \alpha}, \Mat{M}_{\text{part}}\circ \hat{\Mat{I}}_{\text{interp},\alpha}\Big).
\end{equation}
We finetune the converged avatar model together with total loss in \eqnref{eq:total_loss} until it converges.

\subsection{Facial Attribute Transfer from Image}\label{sec:method_lora}
\vspace{-5px}
To transfer facial attribute from an arbitrary image, such as transferring an unseen hairstyle to the reference individual, we need to find the corresponding latent code in our model. 
Although our model can retrieve the latent code by inputting the CLIP~\cite{hong2022avatarclip} features of an image into our MLP as described in \eqnref{eq:latent_mapping_mlp}, it struggles with perfectly handling unseen attributes.
To incorporate these unseen attributes while preserving learned ones, we finetune our avatar model by optimizing the weights $\Delta\Theta$ of additional LoRA~\cite{hu2021lora} layers while keeping the other network weights $\Theta$ frozen. Specifically, our model with additional LoRA layers takes the same inputs and outputs as described in \eqnref{eq:avatar_model}:
\begin{equation}\label{eq:avatar_lora}
    \{\vec{x}_i^{d}, \vec{r}_i^{d}, \vec{o}_i^{d}, \vec{s}_i^{d}, \vec{c}_i^{d} \} = \mathcal{M}_{\Theta+\Delta\Theta}(\mathbf{x}_i^{gc}, \mathbf{z}, \vec{\beta}, \vec{\theta}, \vec{\psi}).
\end{equation}
We animate the image for transfer with our animation generation pipeline and use the resulting frames to optimize the LoRA layers.
The loss is calculated only on the region targeted for transfer, using a masked loss similar to \eqnref{eq:interpolation_loss}. 
Refer to the Supp. Mat. for more details.

\section{Experiments}
\vspace{-2px}
\subsection{Synthetic Dataset Configuration}
\vspace{-3px}
To assess the effectiveness of our method, we generate a synthetic dataset using a single portrait for model evaluation. 
We define 9 attribute categories 
(beard, clothes, earrings, eyebrows, hair, hat, headphones, mouth, and nose)
and produce over 50 videos for each, 
resulting in a total of 957 attribute-edited videos for quantitative comparison.
The text prompts are constructed from non-contradictory combinations of predefined, category-specific adjectives, such as curly, straight, wavy, and coily for hair. 
To animate the images, we employ a single 513-frame video that captures a variety of head poses and expressions, applying it consistently across all instances.
We split all video frames in our synthetic dataset into training and test sets with a 400:113 frame ratio, using the first 400 frames for training and the remaining 113 for evaluation.
Examples of our dataset can be found in \figref{fig:synthetic_dataset}. 

\begin{figure}[t]
\includegraphics[trim={0 0 0 0},clip,width=\columnwidth]{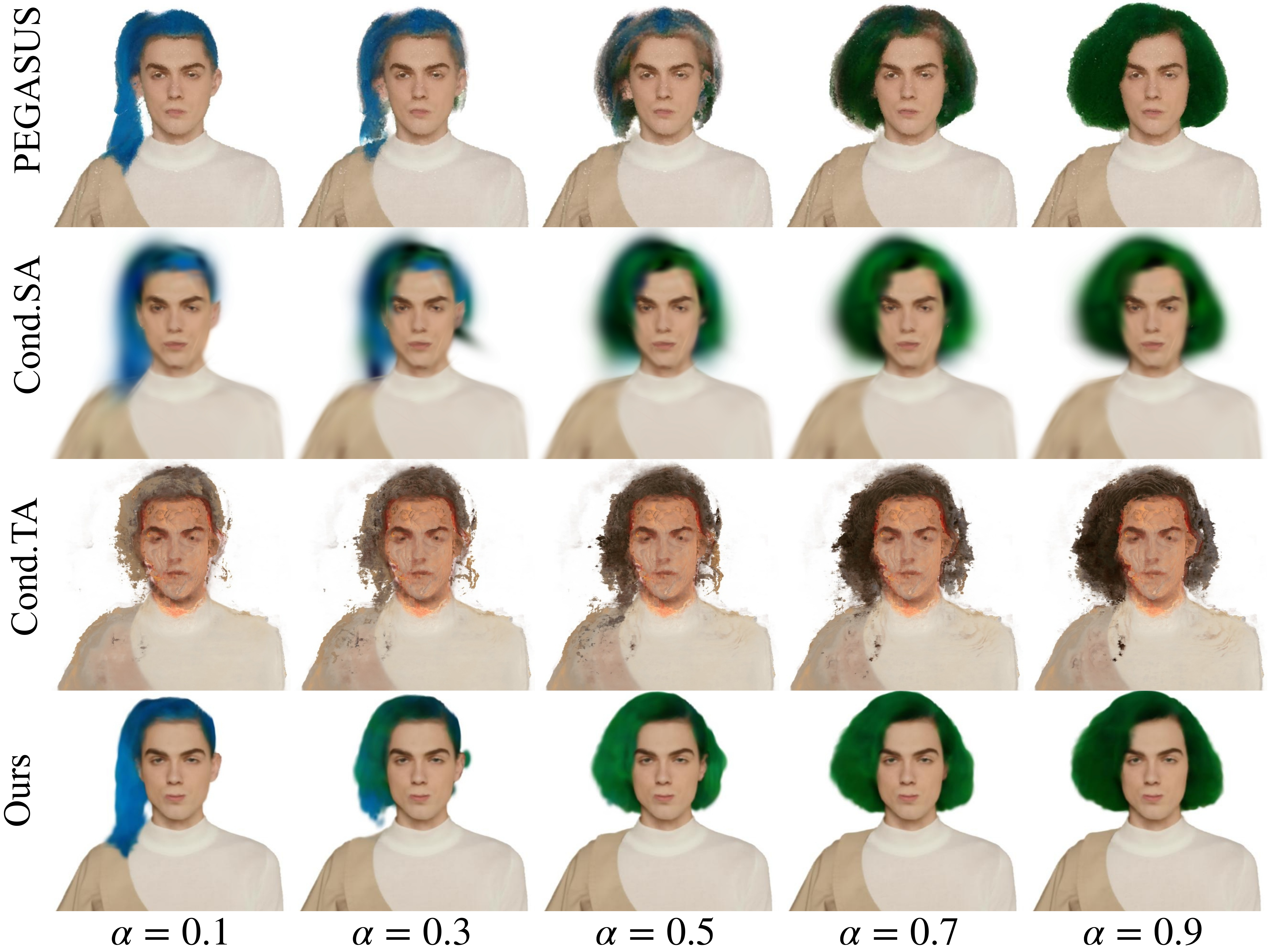}
\vspace{-20px}
\caption{
\textbf{Interpolation Comparison on Baselines.} Our method shows better interpolation smoothness and less artifact on interpolated samples, particularly on the texture and color of hair.
}
\label{fig:qual_baselines}
\vspace{-15px}
\end{figure}    

\subsection{Baselines and Metrics} 
\vspace{-2px}
We compare our model with three different baselines, each using a distinct 3D representation for avatar modeling: colorized point clouds~\cite{zheng2023pointavatar}, NeRF~\cite{mildenhall2021nerf}, and 3D Gaussians~\cite{kerbl20233d}.

\noindent\textbf{PEGASUS}~\cite{cha2024pegasus} is the first method for constructing a personalized 3D generative avatar from 2D monocular video inputs. It creates a personalized avatar model using a set of MLP networks and a colorized point cloud, following the approach of PointAvatar~\cite{zheng2023pointavatar}. For a fair comparison, we train PEGASUS with its public code, replacing its synthetic database with our synthetic datasets.

\noindent\textbf{Conditional INSTA} \emph{(Cond.TA)} is a modified version of vanilla INSTA~\cite{zielonka2023instant}, which reconstructs head avatars using an implicit representation, specifically iNGP~\cite{muller2022instant}. To enable the model to capture diverse facial attributes, we add latent code conditioning the MLP of vanilla INSTA. We follow the PEGASUS latent code configuration and train \emph{Cond.TA} with our synthetic dataset until it converges.

\noindent\textbf{Conditional SplattingAvatar} \emph{(Cond.SA)} is a modified SplattingAvatar~\cite{shao2024splattingavatar} which is a method for reconstructing 3D avatar models from monocular video using 3D Gaussian Splatting~\cite{kerbl20233d}. Vanilla SplattingAvatar explicitly represents an avatar as a set of 3D Gaussians embedded on a 3D head mesh. To incorporate conditional latent code as input, we add an implicit network estimating changes of the 3D Gaussian parameters conditioned by the latent code. Similar to other baselines, we train the model until convergence using our synthetic dataset. See the Supp. Mat. for more details.

\noindent\textbf{Metrics.} 
We evaluate our personalized generative model in two aspects: reconstruction performance and generative performance. 
Following standard practices in monocular 3D avatar reconstruction~\cite{zheng2023pointavatar,shao2024splattingavatar,zielonka2023instant}, we use peak signal-to-noise ratio (PSNR), structural similarity (SSIM), and perceptual similarity (LPIPS)~\cite{zhang2018perceptual} to evaluate reconstruction performance of learned subjects in synthetic dataset.
Additionally, we evaluate identity preservation by computing the cosine similarity of ArcFace~\cite{deng2019arcface} identity features.

We compute the Fréchet Inception Distance (FID)~\cite{heusel2017gans_fid_paper} and Kernel Inception Distance (KID) against FFHQ dataset~\cite{karras2019style} and our synthetic evaluation dataset to assess the quality of generated subjects. 
In addition, we compute the sum (Perceptual Path Length, PPL) and deviation  (Perceptual Distance Variance, PDV) of perceptual loss between adjacent interpolated images to evaluate the smoothness of interpolation following DiffMorpher~\cite{zhang2024diffmorpher, shoemake1985animating}.
\vspace{-2px}

\begin{figure}[t]
\includegraphics[trim={0 0 0 0},clip,width=\columnwidth]{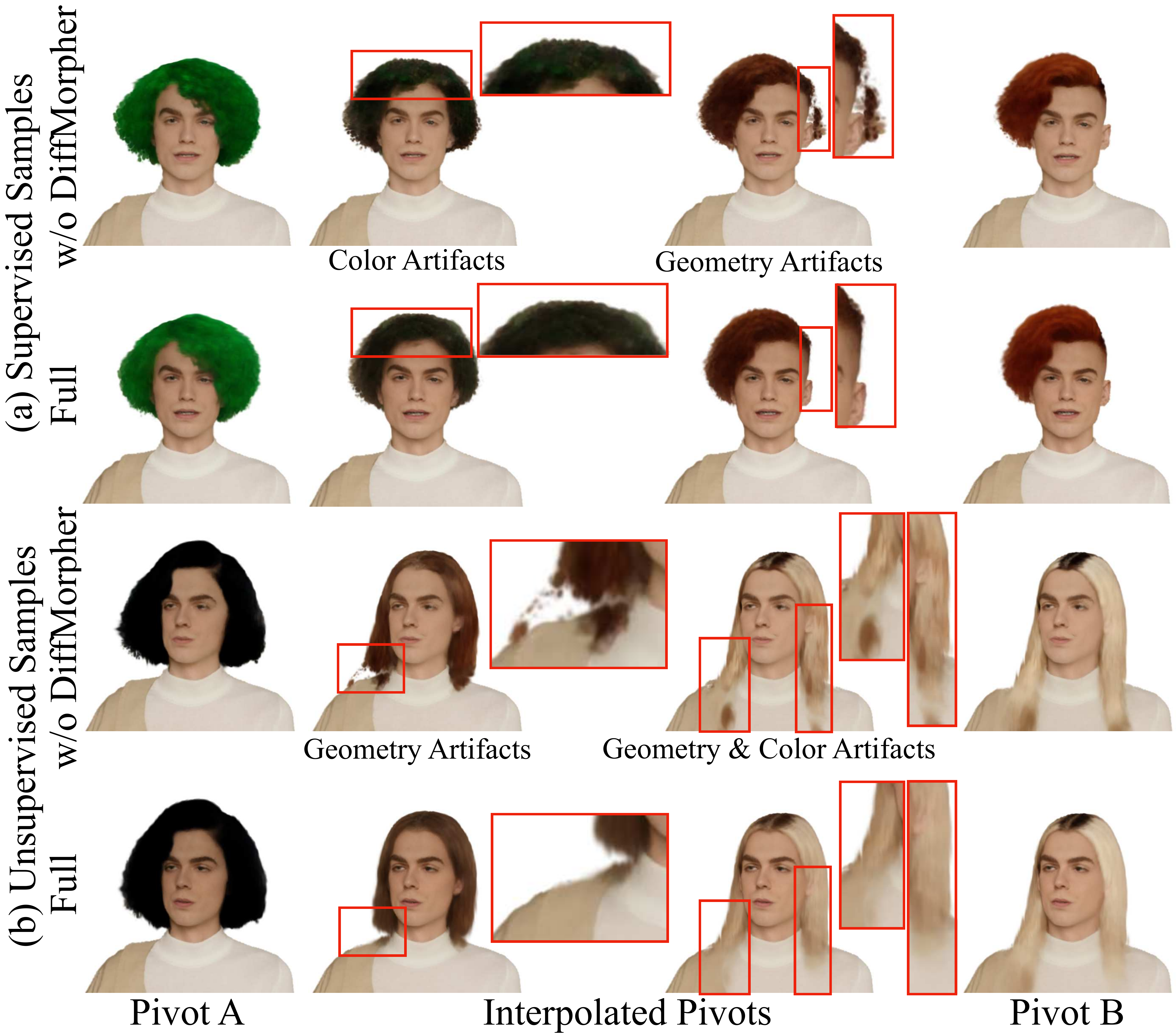}
\vspace{-20px}
\caption{\textbf{Effect of Interpolation Loss.} (a) and (b) represent supervised and unsupervised samples respectively supervised by a personalized diffusion model~\cite{zhang2024diffmorpher, hu2021lora}. Even for unsupervised samples, our supervision method for interpolated samples mitigates unnatural artifacts and textures. Additionally, our method preserves the quality of the pivot samples.}
\label{fig:results_interpolation}
\vspace{-10px}
\end{figure}

\subsection{Quantitative and Qualitative Results}
\vspace{-3px}
 We present the quantitative results of unseen head pose and facial expression rendering in \tabref{tab:quant_avatar_network}. 
 As shown in the table, our avatar model achieves the best results across all metrics, demonstrating superior reconstruction quality for the subjects in the synthetic dataset while preserving the identity.

In \tabref{tab:quant_generative_performance}, we provide additional quantitative comparisons on interpolation, along with qualitative comparisons in \figref{fig:qual_baselines}. Our avatar model outperforms baselines on both FID$_\text{FFHQ}$ and KID$_\text{FFHQ}$ scores, indicating that our interpolated samples align more closely with real human distribution in the FFHQ dataset. Additionally, our model achieves better FID$_\text{SYN}$ and KID$_\text{SYN}$ scores, confirming that our interpolated samples preserve the identity of the reference individual more effectively than the baselines.

While PEGASUS~\cite{cha2024pegasus} achieves slightly better performance on the PDV metric with a small gap, its lower FID, KID, and PPL scores suggest limited naturalness and smoothness in interpolation. 
It can be checked in \figref{fig:qual_baselines}, where PEGASUS shows unnatural transitions in hair color and texture, while ours produces smoother results.
Moreover, in user studies, our interpolation results are preferred over PEGASUS.
\begin{table}[t]
\centering
\small{
\resizebox{0.75\columnwidth}{!}{
\begin{tabular}{l|ccc|c}
\toprule
Method & \thead{PSNR${\uparrow}$} & \thead{SSIM${\uparrow}$} & \thead{LPIPS${\downarrow}$} & \thead{Identity${\uparrow}$} \\ 
\midrule
PEGASUS~\cite{cha2024pegasus} & 23.56 & 0.8661 & 0.1508 & 0.6471 \\
Cond.TA~\cite{zielonka2023instant} & 19.01 & 0.7730 & 0.2875 & 0.3022 \\
Cond.SA~\cite{shao2024splattingavatar} & 22.17 & 0.8690 & 0.2760 & 0.4759 \\
\midrule
Ours & \bf{23.84} & \bf{0.8852} & \bf{0.1458} & \bf{0.7059} \\
\bottomrule
\end{tabular}
}
}
\vspace{-5px}
\caption{\textbf{Quantitative Results of Unseen Pose Renderings.} 
We compare our method with the baselines for training accuracy of pivots in our synthetic dataset. 
Our method achieves the best results across all metrics, demonstrating superior accuracy in reconstructing samples in our synthetic dataset while preserving identity.
}
\vspace{-5px}
\label{tab:quant_avatar_network}
\end{table}

\begin{table}[t]
\centering
\small{
\resizebox{\columnwidth}{!}{
\begin{tabular}{l|cccc|cc|c}
\toprule
Method & \thead{FID$_\text{FFHQ}{\downarrow}$} & \thead{KID$_\text{FFHQ}{\downarrow}$} & \thead{FID$_\text{syn}{\downarrow}$} & \thead{KID$_\text{syn}{\downarrow}$} & \thead{PDV$^*$${\downarrow}$} & \thead{PPL${\downarrow}$} & \thead{User Study${\uparrow}$} \\ 
\midrule
PEGASUS~\cite{cha2024pegasus} & 223.86 & 0.2502 & 84.94 & 0.0959 & \bf{0.2373} & 0.4047 & 36.3\\
Cond.TA~\cite{zielonka2023instant} & 258.48 & 0.3015 & 127.45 & 0.1454 & 0.9724 & 0.6739 & - \\
Cond.SA~\cite{shao2024splattingavatar} & 230.21 & 0.2551 & 180.79 & 0.2316 & 0.9641 & 0.5789 & -\\
\midrule
Ours & \bf{214.46} & \bf{0.2201} & \bf{57.78} & \bf{0.0420} & 0.2481 & \bf{0.3308} & \bf{63.7}\\
\bottomrule
\end{tabular}
}
}
\vspace{-5px}
\caption{\textbf{Quantitative Results of Interpolated Renderings.} 
PDV*=$100\times\text{PDV}$.
Ours shows the best score among the baselines including user study except for PDV.
}
\vspace{-5px}
\label{tab:quant_generative_performance}
\end{table}

\begin{table}[t]
    \small
    \centering
    \resizebox{\linewidth}{!}{
        \begin{tabular}{cc|cccc|cc|c}
            \toprule
            \multicolumn{2}{c|}{Method} & \multicolumn{7}{c}{Interpolation} \\
            \cmidrule(lr){1-2} \cmidrule(lr){3-9} %
              w/ $\mathcal{L}_\text{interp}$ & w/ CLIP & \thead{FID$_\text{FFHQ}{\downarrow}$} & \thead{KID$_\text{FFHQ}{\downarrow}$} & \thead{FID$_\text{syn}{\downarrow}$} & \thead{KID$_\text{syn}{\downarrow}$} & \thead{PDV$^*{\downarrow}$} & \thead{PPL${\downarrow}$} & \thead{Identity${\uparrow}$} \\
            \midrule
             & & 224.00 & 0.2357 & 66.34 & 0.0546 & 0.3046 & 0.3387 & 0.7001 \\
             & \checkmark & 224.67 & 0.2335 & 69.03 & 0.0568 & 0.3037 & \bf{0.3268} & 0.66724 \\
             \checkmark & \checkmark & \bf{214.46} & \bf{0.2201} & \bf{57.78} & \bf{0.0420} & \bf{0.2481} & 0.3308 & \bf{0.7013} \\
        \bottomrule
        \end{tabular}
    }
\vspace{-5pt}
\caption{\textbf{Ablation Studies.} 
PDV*=$100\times\text{PDV}$.
``w/ $\mathcal{L}_\text{interp}$'' denotes fine-tuning model with interpolation loss and ``w/ CLIP'' means using latent conditioned on CLIP feature.
Our full method achieves the best results on all metric except for PPL.
}
\vspace{-10px}
\label{tab:ablation_latent_space}
\end{table}

\subsection{Ablation Studies and More Results}
\begin{figure}[t]
\includegraphics[trim={0 0 0 0},clip,width=\columnwidth]{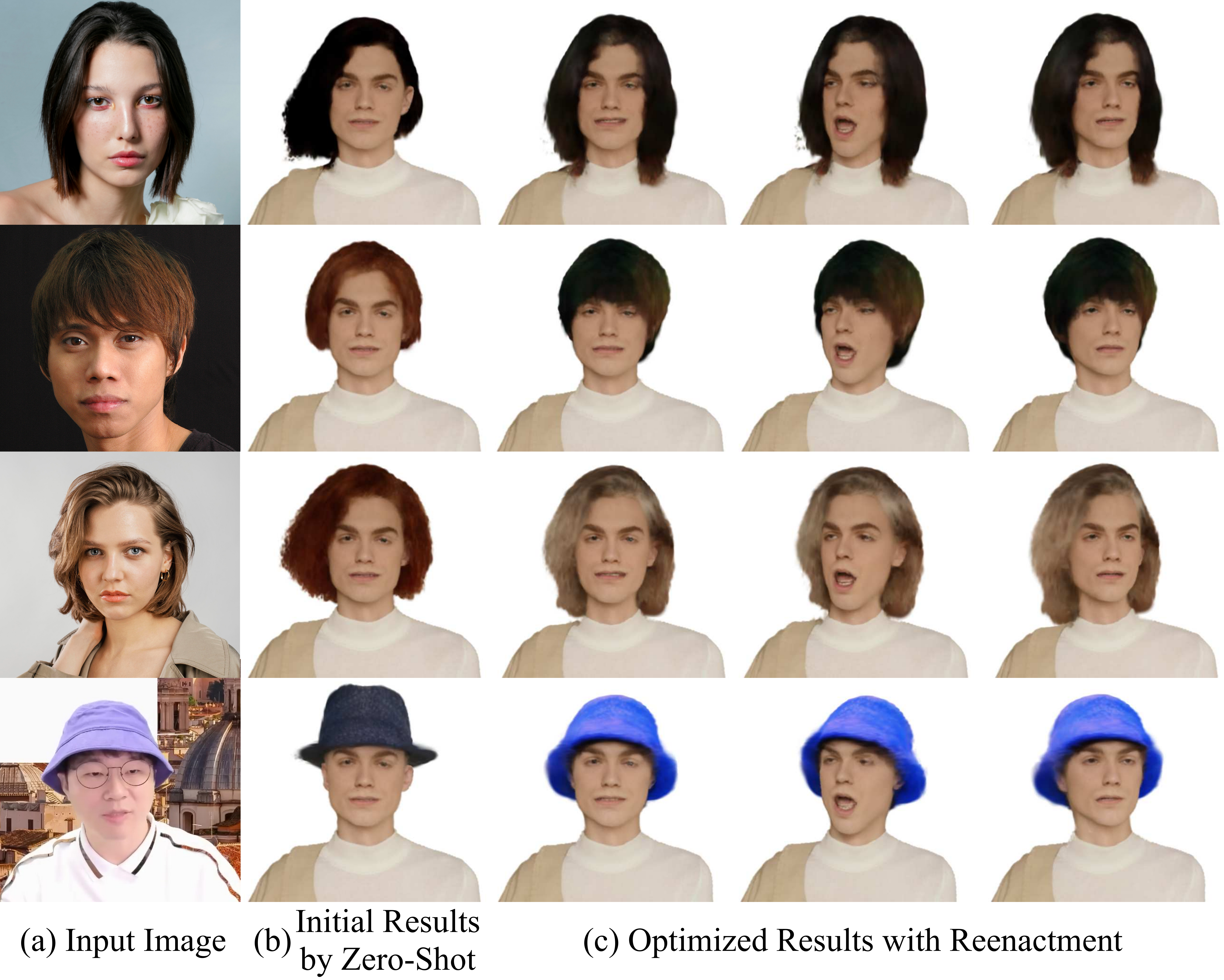}
\vspace{-21px}
\caption{\textbf{Transferred Facial Attribute Results from In-The-Wild Images.} 
(a) is an in-the-wild image of attribute to transfer, (b) is an initial transferred result without optimization, and (c) is optimized results using LoRA layers. 
}
\label{fig:results_lora}
\vspace{-13px}
\end{figure}

\begin{figure}[t]
\includegraphics[trim={0 0 0 0},clip,width=\columnwidth]{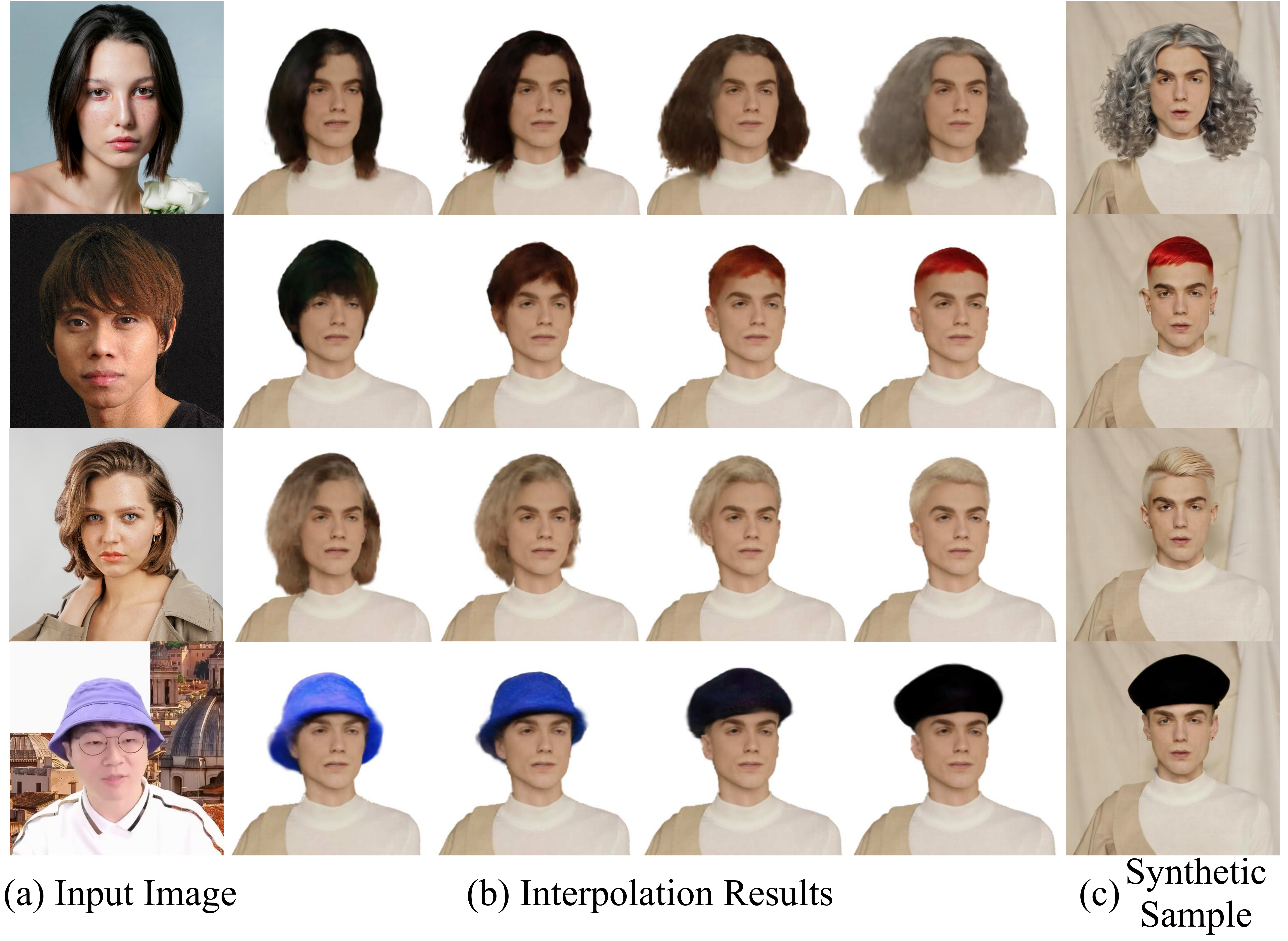}
\vspace{-20px}
\caption{\textbf{Transferred Facial Attribute Interpolation.} 
(a) represents an in-the-wild input image, (b) denotes the interpolation result between (c), a sample of our synthetic dataset.
}
\label{fig:results_lora_interpolation}
\vspace{-20px}
\end{figure}

\noindent\textbf{Ablation Studies.}
We conduct ablation studies to assess the effectiveness of our CLIP-guided latent configuration and interpolation loss $\mathcal{L}_\text{interp}$. As shown in \tabref{tab:ablation_latent_space} and \figref{fig:results_interpolation}, our interpolation loss is essential for improving interpolated sample quality and reducing artifacts. The CLIP-guided latent also reduces PPL, resulting in smoother transitions while preserving rendering quality.

\noindent\textbf{Facial Attribute Transfer.}
We conduct facial attribute transfer experiments using a few in-the-wild images. As shown in \figref{fig:results_lora}, our LoRA fine-tuning method successfully transfers the hair and hat attributes while preserving other aspects of identity. The transferred attributes are well integrated into the latent space, as reflected in the smooth interpolation results between subject in our synthetic dataset in \figref{fig:results_lora_interpolation}

\section{Discussion}
\vspace{-5px}
We present \modelname, an animatable 3D personalized generative avatar from a single portrait image, enabling continuous and disentangled facial attribute editing while preserving the individual's identity.
To achieve this goal, we present several key contributions, including: 
(1) a method to generate high-quality synthetic attribute video datasets from a single image along with our newly trained \champname model;
(2) latent space regularization for unseen or interpolated attribute appearances; and 
(3) an efficient fine-tuning technique via LoRA to integrate new facial attribute into the avatar model.

As limitations, our avatar-building process is computationally intensive, requiring approximately 1.5 days on eight RTX A6000 GPUs for each new identity. Additionally, while our 3D avatars are of high quality, they do not yet achieve photorealism, particularly in fine hair strand details.

\noindent\textbf{Acknowledgments.}
We thank Byungjun Kim for his helpful discussions and advice. This work was supported by NRF grant funded by the Korean government (MSIT) (No. 2022R1A2C2092724), and IITP grant funded by the Korea government (MSIT) [No. RS-2024-00439854, No. RS-2021-II211343, and No.2022-0-00156]. H. Joo is the corresponding author. 

{
    \small
    \setlength{\bibsep}{0pt}
    \bibliographystyle{abbrvnat}
    \bibliography{shortstrings, sections/07_references}
}
\appendix

\clearpage
\newpage

\clearpage

\begingroup
\hypersetup{hidelinks}
\endgroup

\maketitlesupplementary

\section{Implementation Details}
\subsection{Avatar Model}
\subsubsection{Avatar Model Architecture}
To model diverse attributes with a single model, our avatar model follows three-stage deformations proposed in PEGASUS~\cite{cha2024pegasus} with a few modifications. 
First, we initialize the learnable generic canonical points $P^{gc}$ with the vertices of a FLAME~\cite{li2017learning} mesh with an open mouth:
\begin{equation}
    P^{gc} = \{x^{gc}_i\}_{i = \{ 1\cdots N\}},
\end{equation}
where $N$ is the number of points.
The generical canonical points $P^{gc}$ are shared start points for all subjects in the synthetic dataset. 
By deforming the points with subject-specific latent $z$ as a condition, we obtain subject-specific canonical points $P^{sc}$ containing the shape of a specific attribute, such as having long hair or grey cap. 
The mapping between two states is defined as an offset of each point $\mathcal{O}_i^{gc\rightarrow sc}$, which is regressed using coordinate-based deformation MLP as follows:
\begin{equation}\label{eq:deformation_decoder_io}
    \{ \mathcal{O}_i^{gc\rightarrow sc},  \mathcal{O}_i^{sc\rightarrow fc}, \mathcal{E}_i, \mathcal{P}_i, \mathcal{W}_i \} = \text{MLP}_d(\mathbf{z}, \mathbf{x}_i^{gc}).
\end{equation}
It regresses FLAME LBS weight $\mathcal{W}_i$ and blendshapes $\{\mathcal{E}_i,\mathcal{P}_i\}$ of each point jointly, which is crucial to reenact our avatars into any novel pose and expression.
Subsequently, our avatar model defines a mapping of subject-specific canonical points $P^{sc}$ to the FLAME canonical points $P^{fc}$ for better fidelity following the previous work~\cite{zheng2023pointavatar, cha2024pegasus}.
The mappings between two points are defined as another point offset $\mathcal{O}_i^{sc\rightarrow fc}$ which is also regressed by the deforming MLP jointly. The transformation between each state are summarized as follows:
\begin{align}\label{eq:cc_to_sc}
    \mathbf{x}^{sc}_i = \mathbf{x}^{gc}_i + \mathcal{O}_i^{gc\rightarrow sc}, \\
    \label{eq:sc_to_fc}
    \mathbf{x}^{fc}_i = \mathbf{x}^{sc}_i + \mathcal{O}_i^{sc\rightarrow fc}.
\end{align}
Finally, the points in the FLAME-canonical space $P^{fc}$ are deformed into the final posed space $P^d$ using Linear Blend Skinning (LBS) and FLAME parameters $\{\beta,\theta,\psi\}$ as follows: 
\begin{gather}
    \mathbf{x}^{d-} = \mathbf{x}^{fc} + B_{S}(\vec{\beta}; \mathcal{S}) + B_{P}(\vec{\theta}; \mathcal{P}) + B_{E}(\vec{\psi}; \mathcal{E})\\
    \label{eq:lbs}
    \mathbf{x}^{d} = \text{LBS}(\mathbf{x}^{d-}, \mathbf{J}(\psi), \theta, \mathcal{W}),
\end{gather}
where $\mathbf{x}^{d-}$ denotes the point after applying the blendshapes and before applying transformation via linear blend skinning.

\begin{figure}[t]
\includegraphics[trim={0 0 0 0},clip,width=\columnwidth]{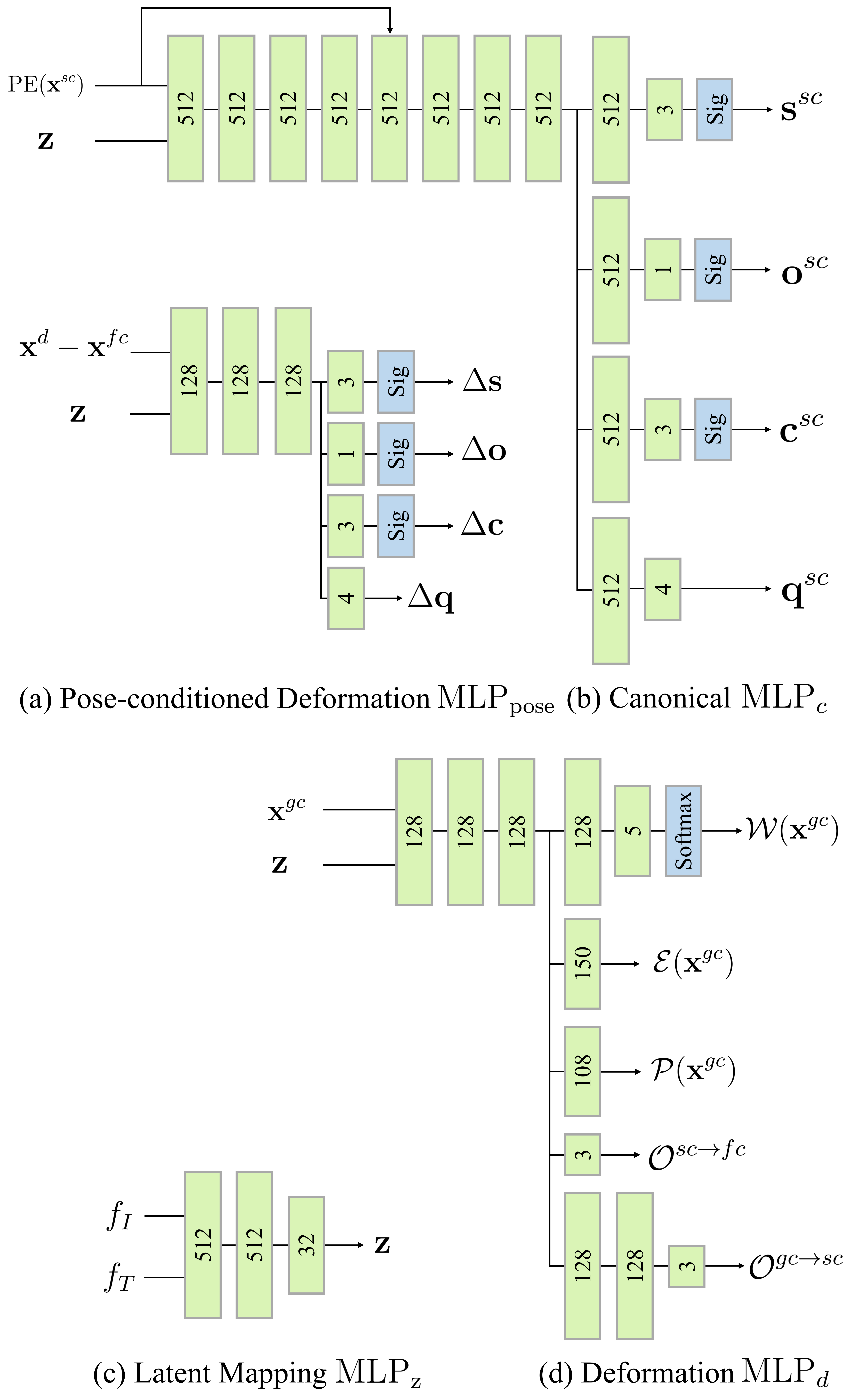}
\vspace{-15px}
\caption{\textbf{Network Configuration.} We show a detailed structure of the networks of our avatar model: pose-conditioned deformation MLP$_\text{pose}$, canonical MLP$_\text{c}$, latent mapping MLP$_\text{z}$, and deformation MLP$_\text{d}$.}
\label{fig:supp_network_configuration}
\vspace{-15px}
\end{figure}

 Similar to PEGASUS~\cite{cha2024pegasus}, we infer the attributes of each Gaussian, $\mathbf{o}_i$ (opacity), $\mathbf{r}_i$ (rotation), $\mathbf{s}_i$ (scale), and $\mathbf{c}_i$ (color) using a coordinated-based MLP as follows:
\begin{equation}\label{eq:canonical_mlp}
    \{ \mathbf{o}_i^{sc}, \mathbf{r}_i^{sc}, \mathbf{s}_i^{sc}, \mathbf{c}_i^{sc} \} = \text{MLP}_c(\mathbf{z}, \mathbf{x}_i^{sc}).
\end{equation}
This canonical MLP$_c$ is defined against subject-specific canonical points and conditioned by latent code $\mathbf{z}$.
 We model additional 3D Gaussian change depending on the pose changes following MonoGaussianAvatar~\cite{chen2024monogaussianavatar}. We calculate the deviation of each Gaussian center between before and after LBS deformation of \eqref{eq:lbs} and query the change of each center to an MLP network together with latent $\mathbf{z}$ to estimate pose-conditioned deformation:
\begin{gather}
    \Delta\mathbf{x}_i = \mathbf{x}^{d}_i - \mathbf{x}^{fc}_i, \\
    \{\Delta \mathbf{r}_i,\Delta \mathbf{s}_i,\Delta \mathbf{o}_i,\Delta \mathbf{c}_i \} = \text{MLP}_{\text{pose}}(\Delta \mathbf{x}_i, \mathbf{z}).
\end{gather}
We change all Gaussian parameters except the center $\mathbf{x}_i$.
The final deformed Gaussians which are queried in the Gaussian Rasterizer~\cite{kerbl20233d} are as follows:
\begin{gather}
    \mathbf{o}_i^d = \Delta\mathbf{o}_i + \mathbf{o}_i^{sc}, \\
    \mathbf{s}_i^d = \Delta\mathbf{s}_i + \mathbf{s}_i^{sc}, \\
    \mathbf{c}_i^d = \Delta\mathbf{c}_i + \mathbf{c}_i^{sc}, \\
    \mathbf{r}_i^d = \Delta\mathbf{r}_i + \text{Rot}(\mathbf{r}_i^{sc}, \frac{\partial \mathbf{x}_i^d}{\partial \mathbf{x}_i^{fc}}),
\end{gather}
where $\text{Rot}(\cdot)$ denotes multiplying a corresponding rotation $\frac{\partial \mathbf{x}_i^d}{\partial \mathbf{x}_i^{fc}}$ on each quaternion $\mathbf{r}_i^{sc}$ occurred during LBS of \eqref{eq:lbs}. 
The overall optimizable parameters of our avatar model are summarized below:
\begin{equation}\label{eq:our_all_avatar}
    \Theta = \{ \text{MLP}_c, \text{MLP}_{d}, \text{MLP}_\text{z}, \text{MLP}_{\text{pose}}, \{\mathbf{x}_i^{gc}\}_{i \in \{1\cdots N\}} \}.
\end{equation}
The detailed network structure is shown in \figref{fig:supp_network_configuration}.

\subsubsection{Training Strategy}
We set the first epoch as a warm-up stage for stable optimization. During this stage, the pose-conditioned deformation MLP is disabled, and only the remaining MLPs and points are optimized. It encourages the deformation module of the avatar network to generate valid offsets from the generic canonical space to the final deformed space. We optimize our avatar model for 112 epochs using DDP with 8 A6000 GPUs, which takes around 2 days. 

We follow prior work~\cite{zheng2023pointavatar, chen2024monogaussianavatar} to iteratively densify the Gaussians via upsampling every 5 epochs until the number of points reaches 130,000. Once this target is reached, we reduce the length of the existing Gaussian attributes’ 3D covariance by a factor of 0.75, and prune Gaussian attributes with opacity lower than 0.5 every 5 epochs. To maintain the point count at 130,000, we additionally upsample new Gaussian attributes with a fixed radius of 0.004.

\subsubsection{Loss Functions} 
The FLAME loss~\cite{zheng2023pointavatar, chen2024monogaussianavatar} included in total loss $\mathcal{L}_{tot}$ is regularization enforcing the inferred FLAME blendshapes and LBS weights $(\hat{\mathcal{E}}, \hat{\mathcal{P}}, \hat{\mathcal{W}})$ of each Gaussian to be close to the FLAME mesh's one:
\begin{gather}
    \begin{split}
        \mathcal{L}_\text{FLAME} = {\frac{1}{N}}&\sum_{i=1}^{N}( \lambda_{e}\|{\mathcal{E}}_{i}-{\hat{\mathcal{E}}}_{i}\|_{2} \\
        & +\lambda_{p}\|\mathcal{P}_{i}-{\hat{\mathcal{P}}}_{i}\|_{2}
        +\lambda_{w}\|\mathcal{W}_{i}-{\hat{\mathcal{W}}}_{i}\|_{2}), 
    \end{split}
\end{gather}
where $\mathcal{E}, \mathcal{P}, ~\text{and} ~\mathcal{W}$ are the pseudo ground truth from the $k$-nearest neighbor vertices of the FLAME~\cite{li2017learning}. This regularization is important to obtain better reenactment with unseen pose.

\subsection{Finetuning for Interpolated Samples}
\subsubsection{Preliminaries: DiffMorpher}
By viewing a diffusion sampling process as a solution of ODE, we obtain a deterministic mapping between a latent variable in the Gaussian distribution $\mathbf{\xi}_T \in \mathcal{N}$ and an image $\Mat{I}$ through DDIM forward and inversion~\cite{song2020denoising}:
\begin{align*}
    \mathbf{\xi} & = \text{DDIM}_\text{inv}(\Mat{I}; \mathbf{W}), \\
    \Mat{I} & = \text{DDIM}(\mathbf{\xi}; \mathbf{W)},
\end{align*}
where $\mathbf{W}$ means a pre-trained image diffusion model. 
By interpolating latents $(\mathbf{\xi}_a, \mathbf{\xi}_b)$ inverted from two images $(\Mat{I}_a, \Mat{I}_b)$, we obtain semantically meaningful smooth interpolation as follows:
\begin{align*}
    \mathbf{\xi}_{interp, \alpha} &= slerp(\mathbf{\xi}_b, \mathbf{\xi}_a, \alpha), \\
    \Mat{I}_{interp, \alpha} &= \text{DDIM}(\mathbf{\xi}_{interp, \alpha}; \mathbf{W}), 
\end{align*}
where $\alpha$ is an interpolation weight and $slerp(\cdot)$ is spherical linear interpolation~\cite{shoemake1985animating}. 

DiffMorpher~\cite{zhang2024diffmorpher} uses personalized diffusion models for DDIM sampling and inversion, resulting in smoother and better natural image interpolation. For two images $(\Mat{I}_a, \Mat{I}_b)$, it trains LoRA~\cite{hu2021lora} on UNet $(\Delta \mathbf{W}_a, \Delta \mathbf{W}_b)$ for each image and uses the LoRA-integrated UNet for DDIM inversion:
\begin{align*}
    \mathbf{\xi}_a & = \text{DDIM}_{inv}(\mathbf{I}_a; \mathbf{W}+\Delta\mathbf{W}_a), \\
    \mathbf{\xi}_b & = \text{DDIM}_{inv}(\mathbf{I}_b; \mathbf{W}+\Delta\mathbf{W}_b).
\end{align*}
For the forward process on interpolated latent $\mathbf{\xi}_{interp, \alpha}$, it uses interpolated LoRA with attention interpolation:
\begin{equation}
    \Mat{I}_{interp, \alpha} = \text{DDIM}(\mathbf{\xi}_{interp, \alpha}; \Theta_{interp, \alpha}) ,
\end{equation}
where $\mathbf{W}_{interp, \alpha}$ is an interpolated LoRA derived as $\mathbf{W}_{interp, \alpha}=\mathbf{W}+\Delta\mathbf{W}_a (1-\alpha)+\Delta\mathbf{W}_b \alpha$. For brevity, we denote the overall interpolation process with DiffMorpher from two images $(\Mat{I}_a, \Mat{I}_b)$ and a weight $\alpha$ as follows:
\begin{equation}
    \Mat{I}_{ \alpha_i} = \text{DiffMorpher}_{\alpha_i}\Big(\Mat{I}_{a}, \Mat{I}_{b}\Big).
\end{equation}

\subsubsection{DiffMorpher LoRA Optimization}
We use DiffMorpher~\cite{zhang2024diffmorpher} to generate interpolated images, which serve as pseudo ground truth to fine-tune our avatar model.
Specifically, we select two subjects from the synthetic dataset and fine-tune the model for interpolated renderings between them.
To obtain the corresponding pseudo ground truth images with DiffMorpher, we require a LoRA for each image.

Training a LoRA for each posed image is computationally prohibitive considering the number of images in our synthetic dataset. Therefore, unlike vanilla DiffMorpher~\cite{zhang2024diffmorpher}, which uses a single image, we train LoRA subject-wise using all animated frames in each subject.
The LoRA training objective is equal to the standard diffusion training objectives~\cite{rombach2022high} as follows:
\begin{align}
    \mathcal{L}(\Delta\Theta) &= \mathbb{E}_{\epsilon, \tau, i}[||\epsilon - \epsilon_{\Theta+\Delta\Theta}(\mathbf{\xi}_{\tau i}, \tau,\mathbf{c}_i) ||^2] , \\
    \mathbf{\xi}_{\tau i} &= \sqrt{\Bar{\alpha}_\tau} \mathbf{\xi}_{0i}+\sqrt{1-\Bar{\alpha}_\tau}\epsilon,
\end{align}
where $\mathbf{\xi}_{0i} = \mathcal{E}(\mathbf{I}_i)$ represents the latent encoded by the VAE encoder of diffusion model, $\mathbf{I}_i$ is the $i^{th}$ animated image of the subject randomly selected at each iteration, $\epsilon\sim\mathcal{N}(0, \mathbf{I})$ is Gaussian noise, and $\mathbf{\xi}_{\tau i}$ denotes the perturbed latent at diffusion step $\tau$. 
To avoid confusion with our model's latent variable $\mathbf{z}$, we use $\mathbf{\xi}$ to refer to the VAE-encoded latents of the diffusion model here.
We train the subject-specific LoRA with batch size 8 for 5 epochs per subject.

\subsubsection{Interpolation Loss Details}
 To enhance the quality of the interpolated sample and ensure interpolation smoothness, we calculate reconstruction loss on the interpolated samples. In every iteration, we randomly sample two subjects $(a,b)$ from the same category of our synthetic dataset, referred to here as pivots. Then, we generate 5 interpolated samples using linear interpolation as follows:
\begin{equation}\label{eq:latent_interpolation}
    \mathbf{z}_{\alpha,i} = \mathbf{z}_a (1-\alpha_i) + \mathbf{z}_b \alpha_i,
\end{equation}
where $\{\alpha_i\}_{[i=1\cdots5]}$ are 5 equally distributed interpolation weights from $1/6$ to $5/6$.
For all 5 interpolated samples, we compare the rendering with DiffMorpher~\cite{zhang2024diffmorpher} generated images as follows:
\begin{gather}
    \hat{\Mat{I}}_{\alpha_i} = \text{GSR}\Big(\mathcal{M}_\Theta(\mathbf{z}_{\alpha,i})\Big), \\
    \Mat{I}_{ \alpha_i} = \text{DiffMorpher}_{\alpha_i}\Big(\Mat{I}_{a}, \Mat{I}_{b}\Big), \\
    \mathcal{L}_\text{interp} = \sum_{i=1}^{5}\mathcal{L}_{\text{part}}\Big(\Mat{M}_{\text{part}}\circ \Mat{I}_{\alpha_i}, \Mat{M}_{\text{part}}\circ \hat{\Mat{I}}_{\alpha_i}\Big).
\end{gather}
As the image $\Mat{I}_{\alpha}$ generated by DiffMorpher~\cite{zhang2024diffmorpher} fails to preserve the identity of the remaining regions, we apply the loss only to the subpart region $M_{\text{part}}$ is that changes during interpolation. 

All DiffMorpher inferences and target part segmentations are performed online during optimization, as the number of possible pairs is too large to process in advance.
We fine-tune our avatar model using an interpolation loss applied to 40 arbitrary pairs per subject, resulting in a total of 38,600 pairs.
In each iteration, we also apply the total loss $\mathcal{L}_{tot}$ to the pivot subjects $(a, b)$ to preserve their quality.

\begin{table}[t]
    \centering
    \resizebox{0.75\columnwidth}{!}{
    \begin{tabular}{c|cc}
    \toprule
 \multicolumn{1}{c}{Category} &  \# of attributes &  w/ \champname \\
    \midrule
        Hair   & 395  & \checkmark  \\
        Beard  & 69     & \checkmark   \\
        Cloth  & 57     & -     \\
        Earrings   & 59   &  \checkmark     \\
        Eyebrows  & 58   & -    \\
        Headphones   & 59   & \checkmark    \\
        Hat & 110 & \checkmark \\
        Mouth  & 75  & -  \\
        Nose & 75 & - \\
    \midrule
        Total & 957 & - \\
    \bottomrule
    \end{tabular}
    }
    \caption{\textbf{Number of Attributes in Our Synthetic Dataset.} We use \champname to animate the portrait images when ‘w/ \champname’ is indicated; otherwise, we use LivePortrait~\cite{guo2024liveportrait}.}
    \label{tab:dataset_configuration}
    \vspace{-10px}
\end{table}

\subsection{Synthetic Dataset}
\subsubsection{Attribute-Edited Portrait Image Generation}
 The number of attributes in each category is shown in \tabref{tab:dataset_configuration}. While we generate approximately $1k$ samples to demonstrate the effectiveness of \modelname, the pipeline can be extended to produce any desired amount, as our synthetic dataset generation process is fully automated. We use FLUX with inpainting controlnet~\cite{flux_controlnet_alpha} for Image-to-Image (I2I) inpainting and SDXL with pose controlnet~\cite{podell2023sdxl,zhang2023adding} for attribute mask generation. 
 
\subsubsection{Training \champname}
Our \champname builds upon the architecture introduced in Champ~\cite{zhu2024champ}, incorporating modifications to enhance 3D-awareness and improve reenactment performance. Specifically, we integrate an additional Variational Autoencoder (VAE) encoder-decoder pair dedicated to normal maps, drawing inspiration from MagicMan~\cite{he2024magicman}. Adopting the dual-branch strategy proposed in MagicMan~\cite{he2024magicman}, we introduce an additional U-Net for the normal maps. This U-Net shares all weights with the original RGB U-Net except for the first layer. The shared layers between the two U-Nets enable cross-domain feature integration, allowing the model to fuse features from both normal map and RGB image. By combining geometric and visual information, our approach enhances the geometric awareness of model, resulting in improved structural coherence.

We replace the original SMPL~\cite{SMPL:2015} rendered motion guidance in vanilla Champ~\cite{zhu2024champ} with FLAME rendering. Specifically, we employ a monocular face capture method~\cite{danvevcek2022emoca} to extract FLAME parameters~\cite{li2017learning}. Using these parameters, we render the FLAME depth map and FLAME normal map. To provide motion guidance for the body, including shoulders, which are not covered by FLAME rendering, we supplement the guidance with full-body keypoints and facial landmarks inferred from RGB videos using DWPose~\cite{yang2023effective}.

We use 5,196 videos from CelebV-Text~\cite{yu2023celebv} datasets to train our \champname. Following previous work~\cite{zhu2024champ}, we train \champname using 8 A6000 GPUs in two stages: 58,732 iterations with a batch size of 32 in stage 1, and 26,450 iterations with a batch size of 8 in stage 2. In stage 1, we optimize the model using randomly sampled frames from videos as an image diffusion model. In stage 2, we train only the temporal motion module with videos while freezing other modules.

\subsubsection{Animating Portrait Images}
Enhancing the reenactment capability of our avatar model requires training videos that cover a wide range of facial expressions and head poses. 
We achieve this by animating portrait images with a motion sequence containing diverse expressions and poses. 
To obtain a motion sequence that satisfies both continuity and the minimal number of frames required by \champname, we record a video for this motion sequence ourselves.

Using a reference portrait image and a predefined motion sequence in an RGB video, we first generate an animated portrait video centered on the reference image using LivePortrait~\cite{guo2024liveportrait}. From this video, we extract normal maps, depth maps, and facial keypoint motion guidance using EMOCA~\cite{danvevcek2022emoca} and DWPose~\cite{yang2023effective}. With this guidance, we animate images edited in the hair, hat, and beard attributes using \champname. For other facial attributes, we directly generate RGB videos using LivePortrait~\cite{guo2024liveportrait}.

\subsection{Attribute Transfer}
To transfer facial attributes from in-the-wild images, we incorporate LoRA layers~\cite{hu2021lora} into the MLP network of the avatar model and optimize these layers. The LoRA layers are trained using animated videos generated from input in-the-wild images. We generate the animated videos following the procedure outlined in Sec. \textcolor{red}{4.2} of the main paper. To ensure only the desired attribute is transferred, we segment the relevant sub-part using an off-the-shelf segmentation network~\cite{khirodkar2025sapiens} and apply a part-wise loss as described in Eq. (\textcolor{red}{15}) of the main paper:
\begin{equation}\label{eq:lora_partwise_loss}
    \mathcal{L}_\text{partwise, lora} = \mathcal{L}_{\text{recon}}\Big(\Mat{M}_{\text{part}}\circ \Mat{I}_\text{itw}, \Mat{M}_{\text{part}}\circ \hat{\Mat{I}}_\text{attr}\Big),
\end{equation}
where $I_\text{itw}$ represents the image from video animated in-the-wild portrait image, $\hat{I}_\text{attr}$ denotes the rendered image with latent $\mathbf{z}_{itw}$ regressed by latent mapping MLP$_\text{z}$ from CLIP features of input in-the-wild image.

We observe that using only the partwise loss fails to preserve reference identity of our avatar model and collapse the pretrained latent space. To address this, we introduce a 3D loss. The 3D loss encourages the LoRA layers in the avatar model to produce the same output as when the LoRA layers are absent. Specifically, Gaussian random latent codes $\mathbf{z}_\text{random}$ from the pretrained latent space are sampled and used as inputs along with the FLAME parameters of an animatable portrait video. The model is trained to minimize the difference between the outputs of the avatar model with and without the LoRA layers, ensuring consistency in 3D Gaussian parameters and 3D positions. Specifically, for the Gaussian attributes inferred with and without LoRA layers:
\begin{align}
    \label{eq:3d_loss_points}
    \{\vec{x}_i^{d}, \vec{r}_i^{d}, \vec{s}_i^{d}, \vec{o}_i^{d},& \vec{c}_i^{d} \}  = \mathcal{M}_{\Theta}(\mathbf{x}_i^{gc}, \mathbf{z}_\text{random}, \vec{\beta}, \vec{\theta}, \vec{\psi}), \\
    \{\vec{x}_{i, \text{lora}}^{d}, \vec{r}_{i, \text{lora}}^{d},& \vec{o}_{i, \text{lora}}^{d}, \vec{s}_{i, \text{lora}}^{d}, \vec{c}_{i, \text{lora}}^{d} \} \notag  \\ & = \mathcal{M}_{\Theta+\Delta\Theta}(\mathbf{x}_i^{gc}, \mathbf{z}_\text{random}, \vec{\beta}, \vec{\theta}, \vec{\psi}),
\end{align}
we calculate the distance between them as follows:
\begin{gather} \label{eq:lora_3d_loss}
     \mathcal{L}_\text{3d} = \|\vec{x}_{i, \text{lora}}^{d} -  \vec{x}_i^{d}\|_1  
    + \|\vec{r}_{i, \text{lora}}^{d} - \vec{r}_i^{d}\|_1 \notag \\
    + \|\vec{o}_{i, \text{lora}}^{d} - \vec{o}_i^{d}\|_1 
    + \|\vec{s}_{i, \text{lora}}^{d} - \vec{s}_i^{d}\|_1
    + \|\vec{c}_{i, \text{lora}}^{d} - \vec{c}_i^{d}\|_1.
\end{gather}
The total loss for LoRA layer optimization is defined as follows:
\begin{equation}\label{eq:total_lora_loss}
    \mathcal{L}_\text{total, lora} = \mathcal{L}_\text{3d} + \mathcal{L}_\text{partwise, lora}
\end{equation}
We perform LoRA layer optimization with a learning rate of $1e^{-4}$ for 5 epochs. 

\section{Evaluation Details}
\subsection{Baseline Implementation Details}
To demonstrate our pipeline's effectiveness, we evaluated our methods compared to three different methods. 
\subsubsection{PEGASUS}
We train PEGASUS~\cite{cha2024pegasus} with our synthetic dataset using publicly available code, strictly following the settings described in the paper, including the latent space configuration and network configurations. The model is trained using DDP across 8 RTX 6000 GPUs until convergence. After point rendering with PyTorch3D~\cite{ravi2020pytorch3d}, no additional denoising steps are applied. 

\subsubsection{Conditional INSTA (Cond.TA)}
 To train INSTA with multiple subjects, we introduce a latent condition to the density MLP network, referred to as Conditional INSTA (Cond.TA). We adopt the PEGASUS~\cite{cha2024pegasus} latent configuration to achieve similar sub-part disentangled control. Since the original density MLP network of INSTA is too small to encode a thousand of attributes, we increase the MLP width from 64 to 512 and the depth from 2 to 4. As this adjustment sacrifices rendering speed and increases training time, we focus our comparisons solely on quality, excluding rendering speed. The final Conditional INSTA model is trained using DDP with 8 RTX 4090 GPUs until convergence. 
 
\subsubsection{Conditional SplattingAvatar (Cond.SA)}
Since SplattingAvatar~\cite{shao2024splattingavatar} does not include any network for receiving conditioning, we incorporate an MLP to deform a single set of shared canonical 3D Gaussians into subject-specific canonical 3D Gaussians, similar to the approach in PEGASUS~\cite{cha2024pegasus}. To ensure a fair comparison, we configure the MLP with the same size as PEGASUS's canonical MLP, providing sufficient capacity to represent all subjects in the synthetic dataset. The densification interval is increased from vanilla SplattingAvatar~\cite{shao2024splattingavatar} to address the low stability of optimization in early stages. Densification is halted after 5 epochs, as the gathered gradients do not converge, possibly due to exposure to different subjects in each iteration. We adopt the same latent configuration as the PEGASUS model, and the final Conditional SplattingAvatar model is trained using DDP on 8 RTX 4090 GPUs until convergence.

\subsection{Interpolation Evaluation Details}\label{sec:interpolation_evaluation_details}
To evaluate the rendering quality of avatars with unseen attributes and interpolation smoothness, we sample avatars from our model using interpolated latent codes. For each of the 9 categories in our synthetic dataset, we randomly select 200 subject pairs and generate 9 interpolated latent codes per pair, following \eqref{eq:latent_interpolation}. The intervals between the sampled latent codes are evenly spaced. Each interpolated latent code is used to render the corresponding avatar in 5 different poses. This process produces 9,000 images per category and a total of 81,000 images across all categories for evaluation.

\paragraph{Metrics.} We compute FID and KID scores by comparing our renderings with two different image sets: FFHQ~\cite{karras2019style} and our synthetic evaluation dataset, which is built with the same input reference individual.  
 Specifically, we use $(\text{FID}_\text{FFHQ},\text{KID}_\text{FFHQ})$ to asses the realism and quality of the renderings by comparing with real face images, and $(\text{FID}_\text{SYN}, \text{KID}_\text{SYN})$ to evaluate identity preservation by comparison with the synthetic evaluation dataset. 
 
Since the rendered outputs do not include backgrounds, we remove the backgrounds of all portrait images in FFHQ using MODNet~\cite{MODNet} before calculating metrics. The synthetic evaluation image sets are constructed with the same reference image, following our edited portraits generation pipeline. 
To prevent potential information leaks, we synthesize $2k$ novel images using text prompts not included in the training dataset.
This approach provides a more reliable measurement of identity preservation during attribute editing, particularly for changes that partially alter identity features, such as the eyes, eyebrows, and nose, which are challenging to evaluate with existing identity metrics like ArcFace~\cite{deng2019arcface}.

\subsection{User Studies}

\begin{figure*}[t]
\centering
\includegraphics[trim={0 0 0 0},clip,width=\textwidth]{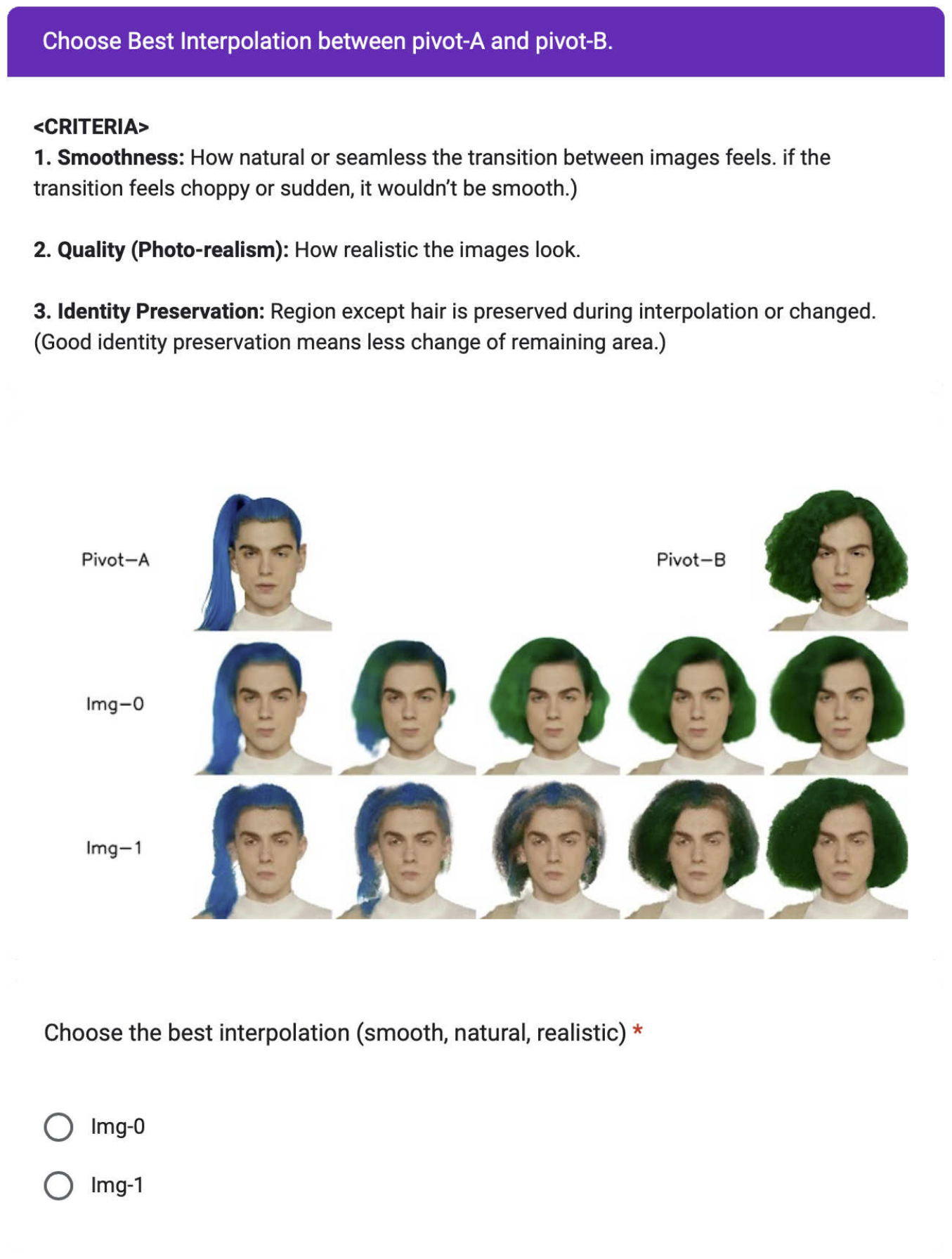}
\vspace{-1.5em}
\caption{\textbf{User Study.} We show user study screenshot.}
\label{fig:supp_user_study}
\vspace{-1.5em}
\end{figure*}

We also conduct a user study to evaluate the rendering quality of interpolated samples, as shown in \figref{fig:supp_user_study}. Since only PEGASUS~\cite{cha2024pegasus} and our method receive votes among the four methods in preliminary study, we exclude \emph{Cond.TA} and \emph{Cond.SA} from the options. Participants are asked to choose the better images based on interpolation smoothness and image quality for 20 pairs of interpolations. The pairs are randomly selected from the hair category. We collect responses from 229 participants via CloudResearch~\cite{cloudresearch}. 

\begin{table}[t]
\vspace{-5px}
\centering
\resizebox{\columnwidth}{!}{
\begin{tabular}{l|ccc|cc}
\toprule
\multirow{2.5}{*}{Method} & \multicolumn{3}{c|}{Accuracy} & \multicolumn{2}{c}{Naturalness} \\
\cmidrule(lr){2-4} \cmidrule(lr){5-6}
& PSNR$\uparrow$ & SSIM$\uparrow$ & LPIPS$\downarrow$ & FID$\downarrow$ & KID$\downarrow$ \\
\midrule
Moore-AnimateAnyone~\cite{hu2024animate, MooreThreads2023MooreAnimateAnyone} & 17.77 & 0.6841 & \emph{0.2536} & \bf{146.59} & \bf{0.0530} \\
MimicMotion~\cite{zhang2024mimicmotion} & 17.27 & 0.6641 & 0.3012 & 178.87 & 0.0980 \\
MegActor-$\Sigma$~\cite{yang2024megactor, yang2024megactorsigma} & \emph{17.89} & \emph{0.6986} & 0.2599 & 155.04 & 0.0572 \\
\midrule
Ours (\champname) & \bf{20.58} & \bf{0.7417} & \bf{0.1878} & \emph{150.59} & \emph{0.0555} \\
\bottomrule
\end{tabular}
}
\vspace{-5px}
\caption{\textbf{Quantitative Comparisons for Image-to-Video Models.} 
We evaluate our \champname with recent diffusion based baselines in face reenactment scenarios. 
Ours \champname obtain the best scores in accuracy and comparable FID and KID.}
\label{tab:quant_image_to_video}
\end{table}

\begin{table}[t]
\centering
\small{
\resizebox{1.0\columnwidth}{!}{
\begin{tabular}{l|c|ccc|c}
\toprule
\multirow{2.5}{*}{Input Type} & \multicolumn{1}{c|}{2D Video} & \multicolumn{4}{c}{Rendering Quality} \\
\cmidrule(lr){2-2} \cmidrule(lr){3-6}
& \thead{Subject Consistency ${\uparrow}$} & \thead{PSNR${\uparrow}$} & \thead{SSIM${\uparrow}$} & \thead{LPIPS${\downarrow}$} & \thead{Imaging Quality${\uparrow}$} 
\\ 
\midrule
Real Video & \bf{0.9761} & \bf{22.26} & 0.9045 & \bf{0.1352} & 0.5366 \\
Synthetic Video~$_\text{self-driving}$ & 0.9719 & 21.23 & \bf{0.9241} & 0.1582 & \bf{0.5896} \\ 
\bottomrule
\end{tabular}
}
}
\vspace{-5px}
\caption{\textbf{Quantitative Comparison of Impact of Inconsistency.} 
Quantitative comparison of PERSE avatar models trained on real and synthetic videos. Note that 2D video evaluated for subject consistency is used for training, and rendering quality is evaluated on unseen head poses and facial expressions using a test sequence.
}
\label{tab:supp_01_consistent_synthetic}
\end{table}

\begin{figure}[t]
\centering
\includegraphics[trim={0 0 0 0},clip,width=\columnwidth]{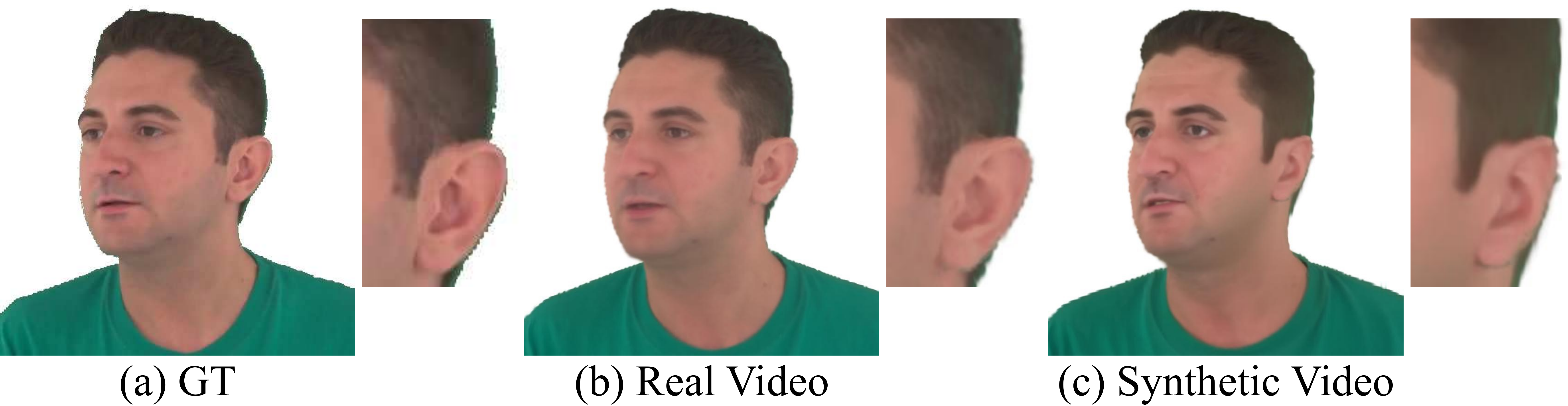}
\vspace{-15px}
\caption{\textbf{Qualitative Comparison of Impact of Inconsistency.} We show qualitative comparision of impact of inconsistency between real and 2D generated video.}
\label{fig:supp_inconsistency}
\end{figure}

\section{More Experiments}
\subsection{Additional Results}
\begin{figure*}[t]
\centering
\includegraphics[trim={0 0 0 0},clip,width=\textwidth]{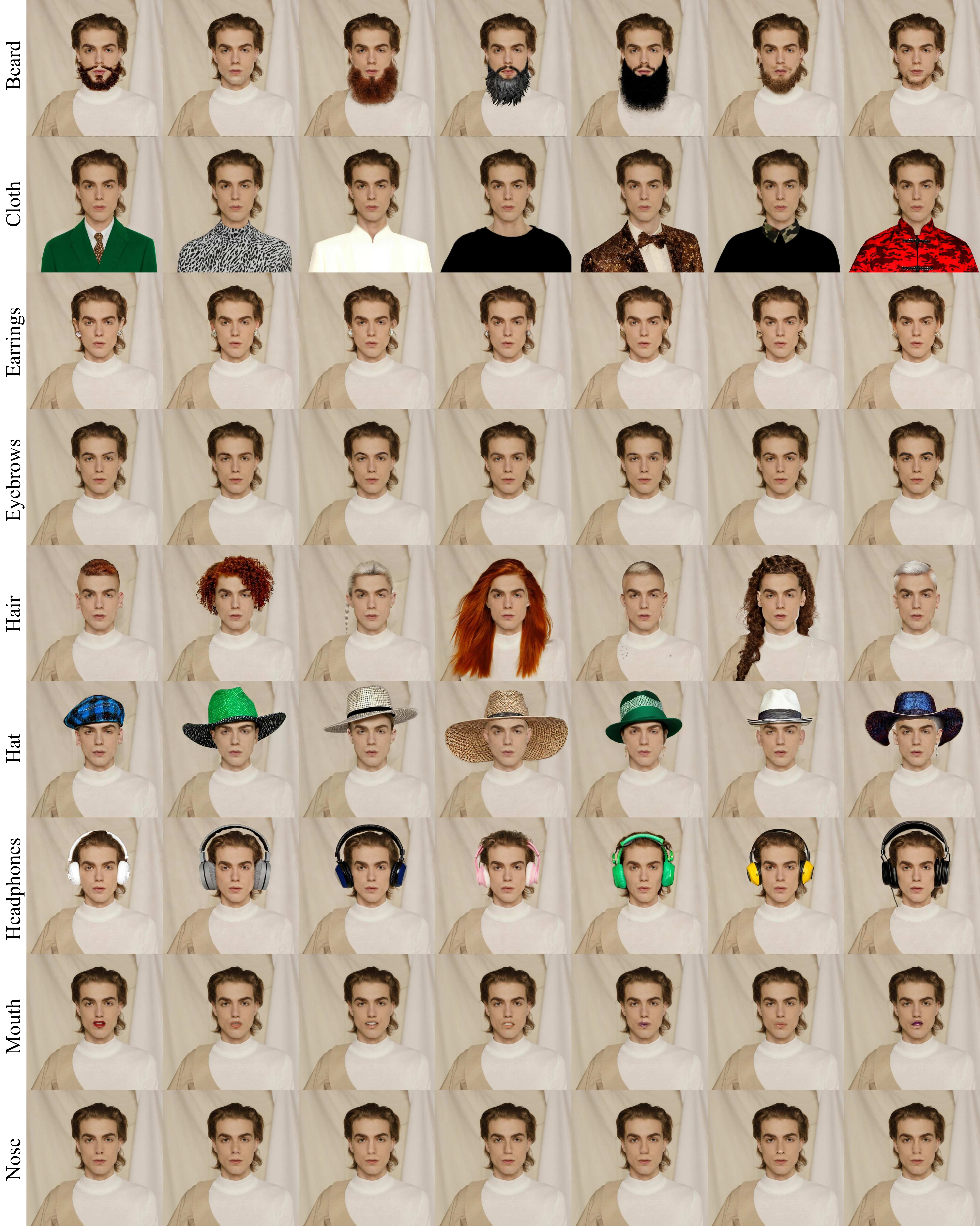}
\vspace{-1.5em}
\caption{\textbf{Example of Attribute-Edited Portrait Image Generation (1).} We present samples of attribute-edited portrait image generation. For each attribute, we display results obtained through random sampling.}
\label{fig:supp_attribute_edited_portrait1}
\vspace{-1.5em}
\end{figure*}

\begin{figure*}[t]
\centering
\includegraphics[trim={0 0 0 0},clip,width=\textwidth]{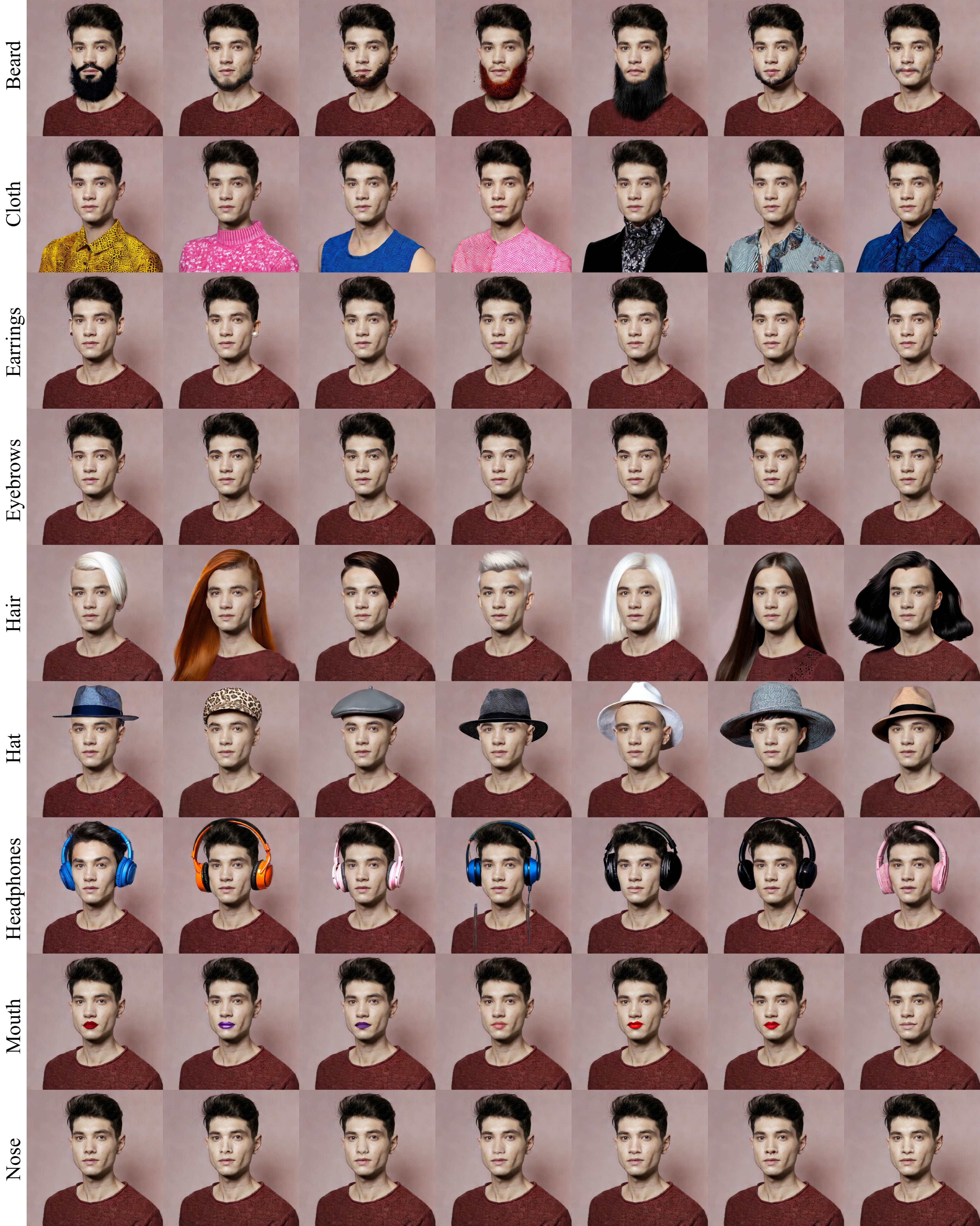}
\vspace{-1.5em}
\caption{\textbf{Example of Attribute-Edited Portrait Image Generation (2).} Our method can be applied to various portrait images}
\label{fig:supp_attribute_edited_portrait2}
\vspace{-1.5em}
\end{figure*}

\begin{figure*}[t]
\centering
\includegraphics[trim={0 0 0 0},clip,width=\textwidth]{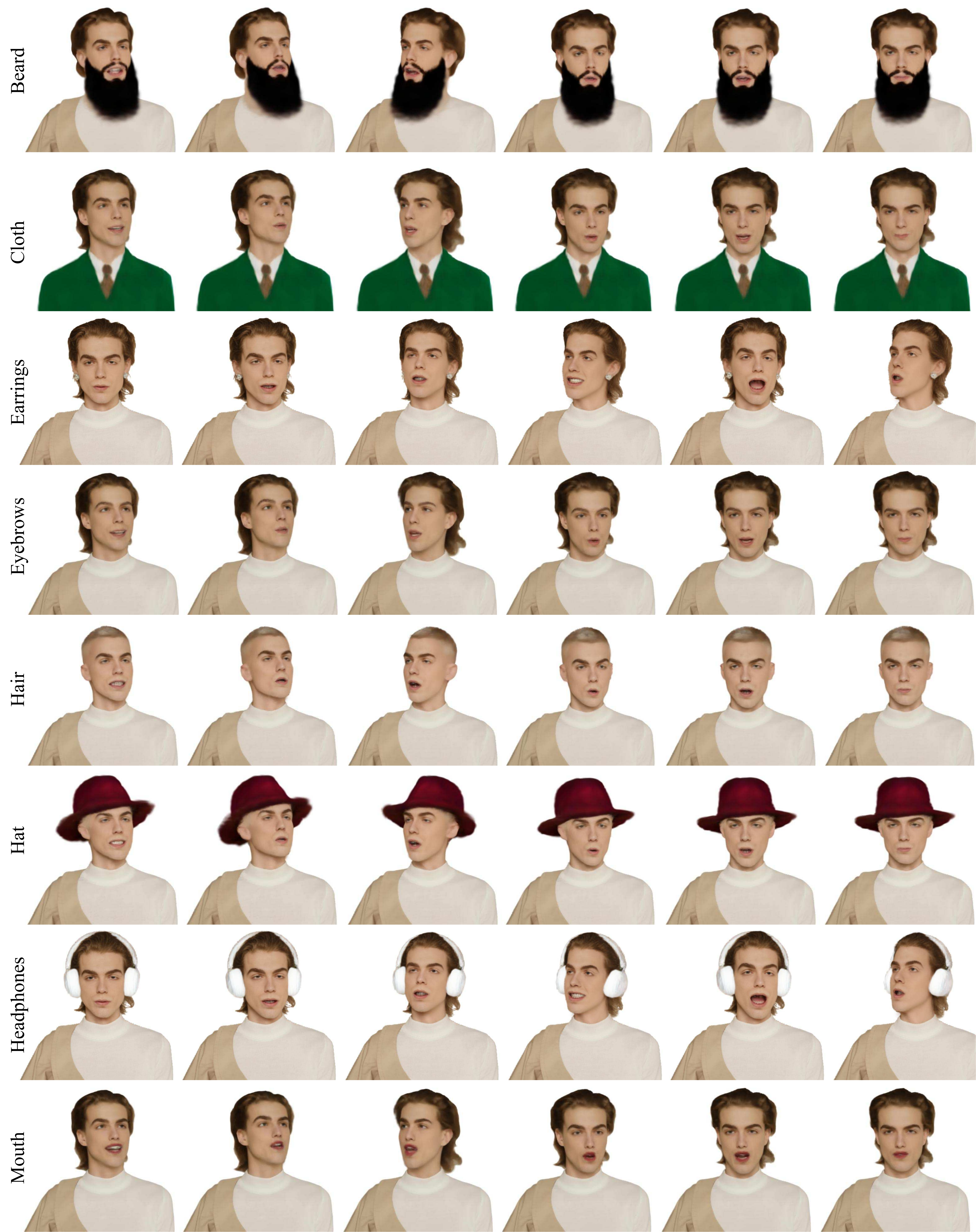}
\vspace{-1.5em}
\caption{\textbf{Unseen Pose Rendering Results (1).} We present the rendering results using latent codes for novel head poses and facial expressions not included in the training dataset, categorized by each attribute.}
\label{fig:supp_unseen_pose1}
\vspace{-1.5em}
\end{figure*}

\begin{figure*}[t]
\centering
\includegraphics[trim={0 0 0 0},clip,width=\textwidth]{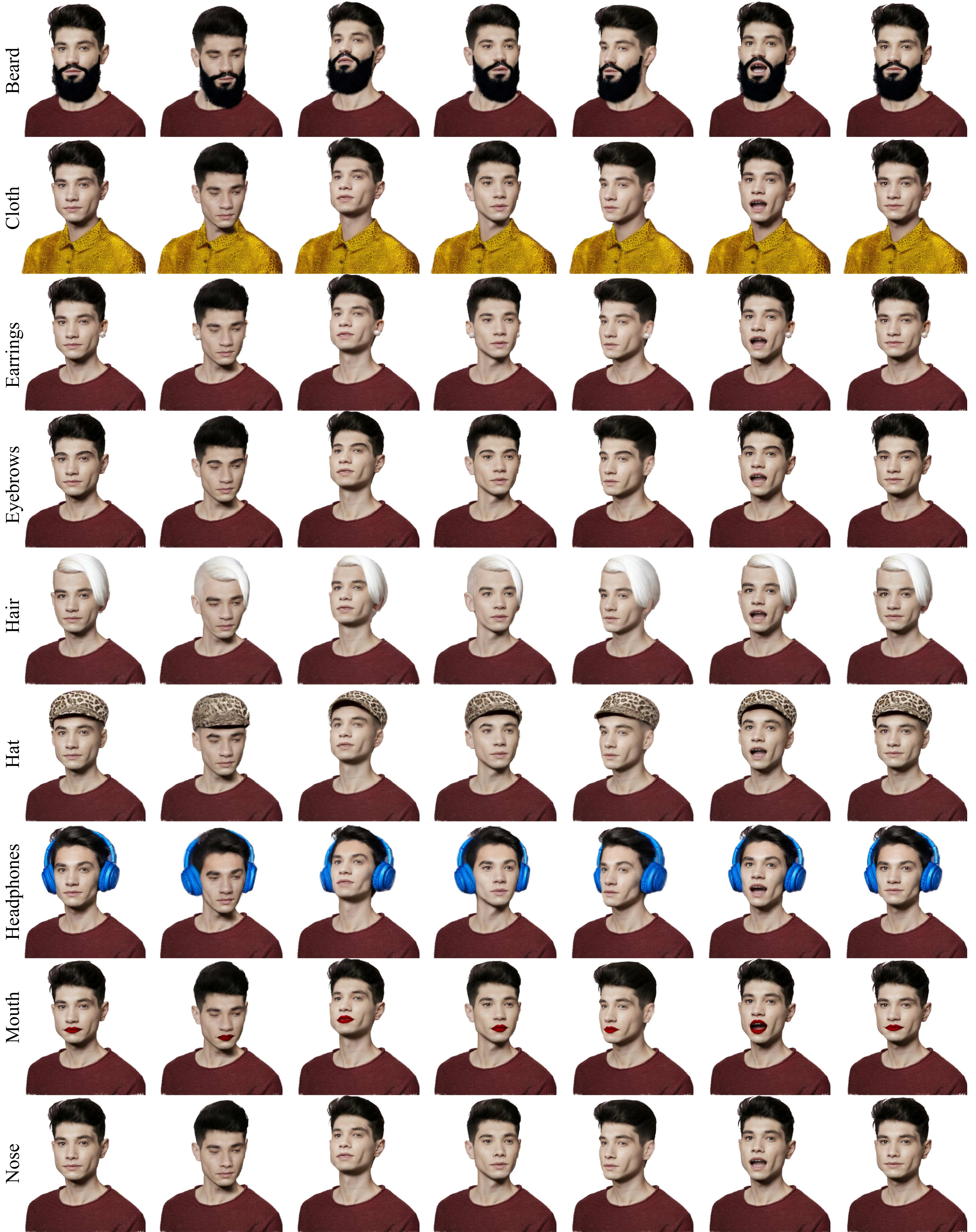}
\vspace{-1.5em}
\caption{\textbf{Unseen Pose Rendering Results (2).}}
\label{fig:supp_unseen_pose2}
\vspace{-1.5em}
\end{figure*}

\begin{figure*}[t]
\centering
\includegraphics[trim={0 0 0 0},clip,width=\textwidth]{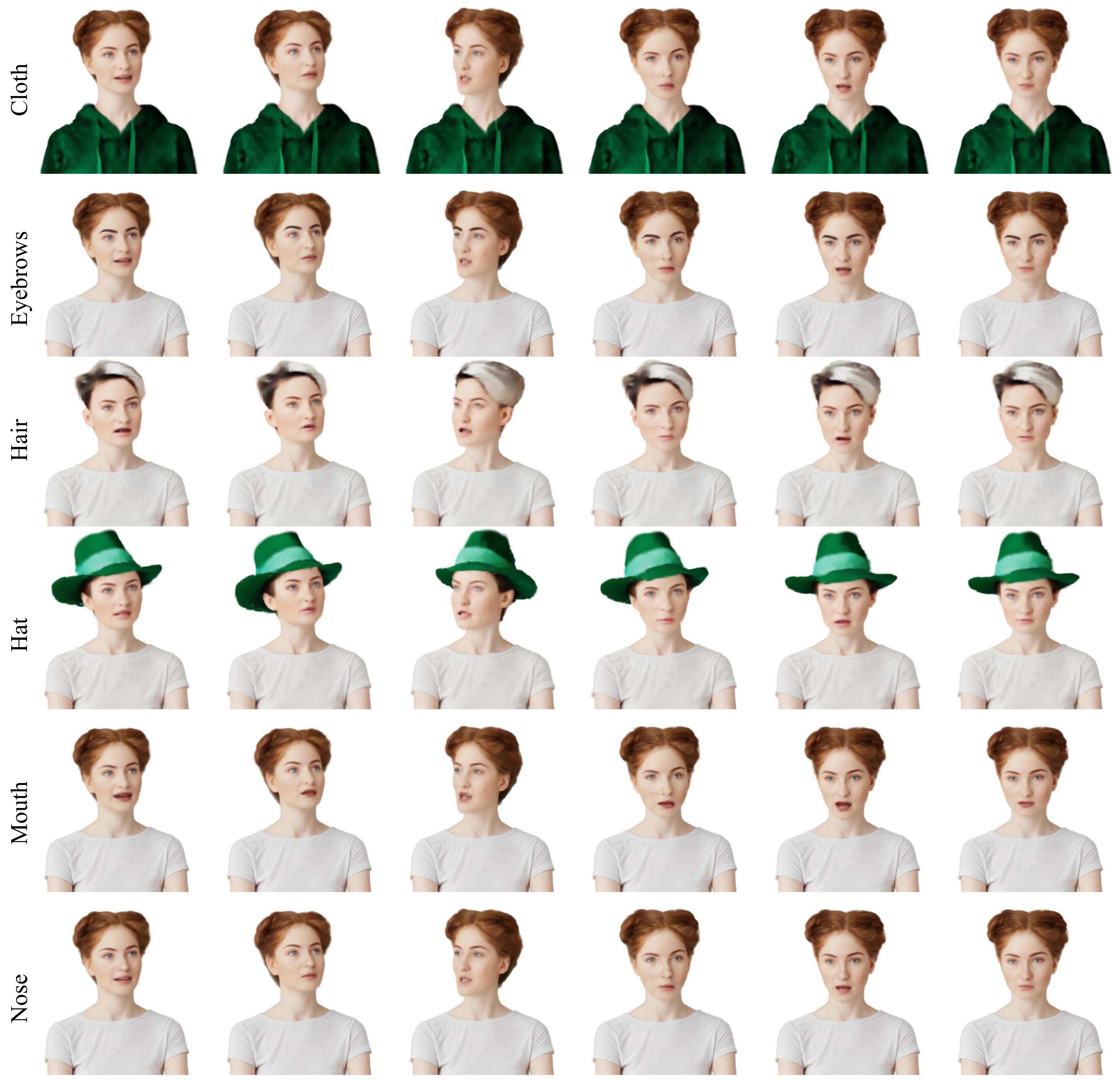}
\vspace{-1.5em}
\caption{\textbf{Unseen Pose Rendering Results (3).}}
\label{fig:supp_unseen_pose3}
\vspace{-1.5em}
\end{figure*}

\begin{figure*}[t]
\centering
\includegraphics[trim={0 0 0 0},clip,width=\textwidth]{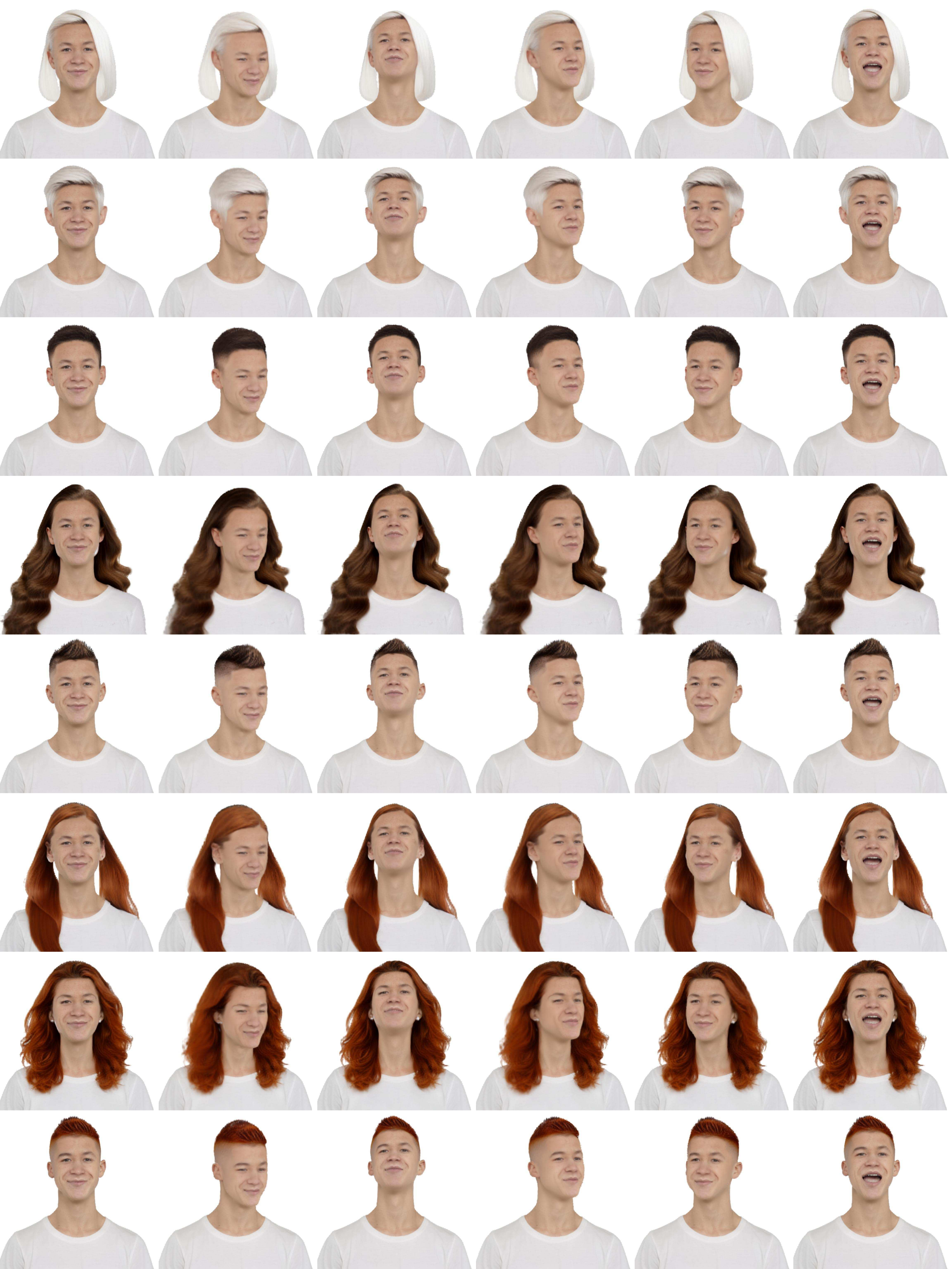}
\vspace{-1.5em}
\caption{\textbf{Unseen Pose Rendering Results (4).} We show hair-only rendering results for unseen poses.}
\label{fig:supp_unseen_pose4}
\vspace{-1.5em}
\end{figure*}

\begin{figure*}[t]
\centering
\includegraphics[trim={0 0 0 0},clip,width=\textwidth]{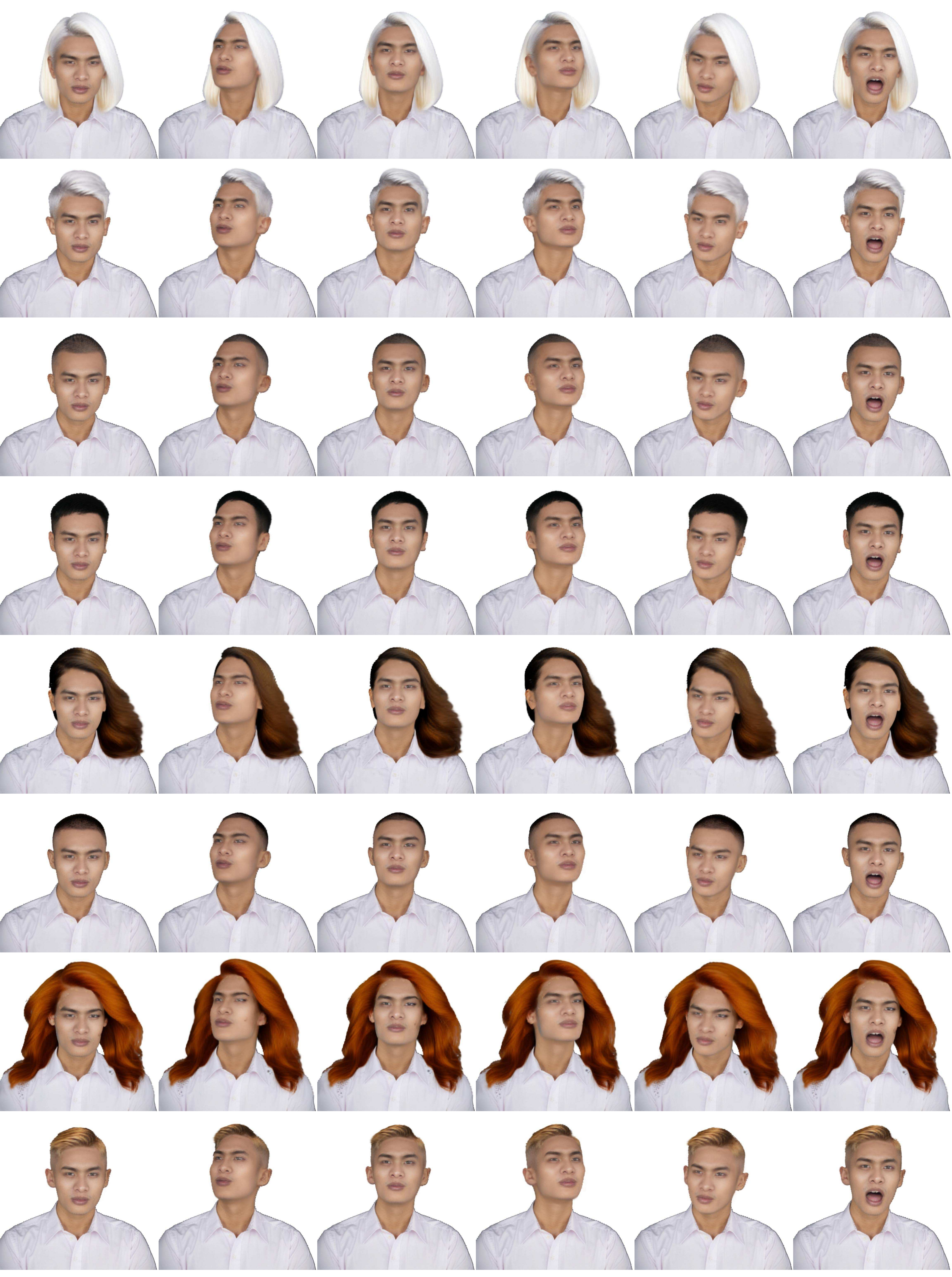}
\vspace{-1.5em}
\caption{\textbf{Unseen Pose Rendering Results (5).} We show hair-only rendering results for unseen poses.}
\label{fig:supp_unseen_pose5}
\vspace{-1.5em}
\end{figure*}

\begin{figure*}[t]
\centering
\includegraphics[trim={0 0 0 0},clip,width=\textwidth]{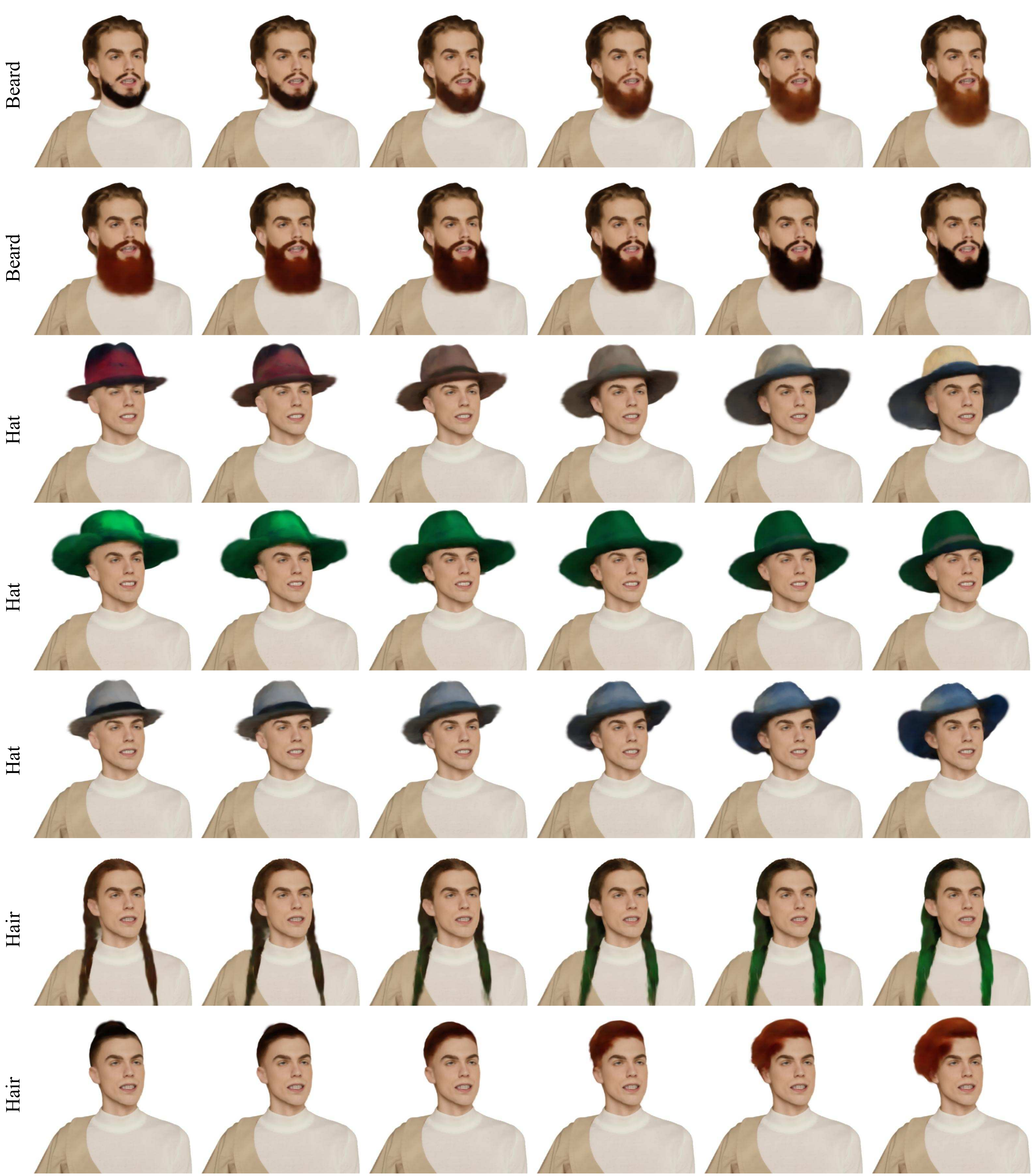}
\vspace{-1.5em}
\caption{\textbf{Interpolation Between Two Latent Codes.} We present the rendering results obtained by interpolating between two latent codes.}
\label{fig:supp_interpolation}
\vspace{-1.5em}
\end{figure*}

We present additional sample results of attribute-edited portrait image generation, providing seven results for each attribute in \figref{fig:supp_attribute_edited_portrait1} and \figref{fig:supp_attribute_edited_portrait2}. Furthermore, we demonstrate the rendering results of our personalized 3D generative avatar on unseen poses, trained with synthetic datasets created using additional portrait images in \figref{fig:supp_unseen_pose1}, \figref{fig:supp_unseen_pose2}, \figref{fig:supp_unseen_pose3}, \figref{fig:supp_unseen_pose4}, and \figref{fig:supp_unseen_pose5}. Finally, we provide the interpolation results between two latent codes for each attribute in \figref{fig:supp_interpolation}.

\subsection{Impact of Video Inconsistency}
The different between real and generated 2D video is negligible, as monocular avatar-building pipeline handles temporal deformations and inconsistencies. To assess this, we present evaluations by building a 3D avatar from each single video, as demonstrated in \tabref{tab:supp_01_consistent_synthetic} and \figref{fig:supp_inconsistency}. We measure subject consistency and imaging quality following VBench~\cite{huang2024vbench}, comparing real video and generated video from \champname by animating the first frame in a self-driven manner, where they show minor differences. After building 3D avatars from each 2D real and generated video separately, we also compare the rendering quality under novel head poses and facial expressions. As shown in \tabref{tab:supp_01_consistent_synthetic} and \figref{fig:supp_inconsistency}, the avatar renderings also show negligible differences in quality, with comparable PSNR, LPIPS, and SSIM scores.

\subsection{Synthetic Monocular Dataset Generation from Single Image}
To demonstrate the effectivness of our \champname, we evaluate the reconstruction quality and rendering realism compared to diffusion based baselines. 
Moore AnimateAnyone~\cite{MooreThreads2023MooreAnimateAnyone} is open-source repository fine-tuned AnymateAnyone~\cite{hu2024animate} to be specialized on facial reenactment. 
MimicMotion~\cite{zhang2024mimicmotion} is a full body animating model based on Stable Video Diffusion~\cite{blattmann2023stable} also capable of reenactment using facial landmarks in DWPose.
MegActor-$\Sigma$ is Diffusion Transformer~\cite{peebles2023dit} based approach to solve reenactement problem. 
We disable the additional audio input option of MegActor-$\Sigma$ during test here. 

We test the methods using 20 sequence randomly selected from CelebV-Text dataset~\cite{yu2023celebv} not seen during the trainining. 
We animate the first frame to make other frames and compare with ground truth frames in the video to compute accuracy. 
We additionally calculate FID and KID against FFHQ dataset~\cite{karras2019style} to evaluate the naturaless of the animated images.
 As shown in \tabref{tab:quant_image_to_video}, our approach achieves the highest reconstruction score across all metrics compared to previous SOTA methods.

\section{Rights}
All portrait reference images used in this work are sourced from the \emph{FreePik}~\cite{freepik_portrait} website under a free license. Note that all of our portraits to show our results are not AI-generated images. Our code and samples of synthetic datasets are publicly released for research purposes only. For more details, refer to \url{https://github.com/snuvclab/perse} about our implementations.

\section{Notations}
Refer to \tabref{tab:notations} for an overview of the notations used in this paper.

\clearpage
\begin{table*}[h]
\caption{Table of notations.}
\label{tab:notations}
\centering
\makebox[0.8\textwidth]{
\begin{tabular}{ll}
\toprule
Symbol  & Description \\
\midrule
\multicolumn{2}{c}{\bf Index}\\
$i$      & Gaussian index $i\in \{1,\dots, N\}$ in 3D Gaussian attributes\\
$j$      & Category index $j\in \{1,\dots, N_c\}$ of edited attributes in synthetic dataset. \\
\midrule
\multicolumn{2}{c}{\bf Learnable Parameters and Networks}\\
MLP$_c$      & Canonical MLP estimating attributes of 3D Gaussians \\
MLP$_d$        & Deformation MLP estimating deformation attributes \\
MLP$_{pose}$        & Pose-conditioned deformation MLP estimating change of Gaussian attributes \\
MLP$_\text{z}$        & Latent mapping MLP from CLIP feature $f_I, f_T$ to subject-specific latent $z$\\
$P^{gc} = \{x^{gc}_i\}_{i = \{ 1\cdots N\}}$ & Learnable positions of 3D Gaussians \\
\midrule
\multicolumn{2}{c}{\bf Spaces of our Avatar Model}\\
$P^{gc}$      & Generic canonical space, single space shared on all subject \\
$P^{sc}$      & Subject-specific canonical space, conditioned by subject latent $\vec{z}$ \\
$P^{fc}$      & FLAME-canonical space, deformed from subject-specific canonical space with blendshape \\
$P^{d}$      & Deformed space, deforming $P^{fc}$ with FLAME pose parameters \\
\midrule
\multicolumn{2}{c}{\bf Diffusion Related}\\
$T$      & Text-prompt queried into the diffusion model \\
$C(\cdot)$  & 2D key points and face landmarks estimator and renderer (OpenPose) \\
$\tau$      & Diffusion denoising time-step \\
$\mathbf{\xi}_0$ & Encoded latent of the queried RGB images of diffusion model\\
$\mathbf{\xi}_\tau$ & Perturbed latent with noise time-step $\tau \in [0,1]$\\
$\epsilon$ & Noise added to the latent\\
\midrule
\multicolumn{2}{c}{\bf Attributes of 3D Gaussians}\\
$\vec{x}_i\in \mathbb{R}^3$& Center of $i$-th Gaussian, or point position in PEGASUS \\
$\vec{q}_i\in \mathbb{R}^4$  & Covariance Matrix's Quaternion of $i$-th Gaussian \\
$\vec{s}_i\in \mathbb{R}^3$  & Covariance Matrix's Scale Component of $i$-th Gaussian \\
$\vec{c}_i\in \mathbb{R}^3$  & Color of $i$-th Gaussian \\
$\vec{o}_i\in \mathbb{R}$  & Opacity of $i$-th Gaussian \\
\midrule
\multicolumn{2}{c}{\bf Off-the-Shelf Network}\\
$\text{I2I}_\text{inpaint}$ & Text-conditioned Image-to-Image inpainting pipeline, based on image diffusion\\
$\text{T2I}$ & Text-to-Image diffusion model\\
$\text{I2V}$ & Portrait animating Image-to-Video model, \champname\ or LivePortrait~\cite{guo2024liveportrait}. \\
\midrule
\multicolumn{2}{c}{\bf FLAME Parameters of Avatar Deformation}\\
$\vec{\theta} \in \mathbb{R}^{15} $ & FLAME pose parameter \\
$\vec{\beta} \in \mathbb{R}^{100} $ & FLAME shape parameters \\
$\vec{\psi} \in \mathbb{R}^{50} $ & FLAME expression parameters \\
$\mathcal{E} \in \mathbb{R}^{50\times 5023}$ & FLAME expression blendshape parameters, estimated by MLP$_d$ for each Gaussian\\
$\mathcal{P} \in \mathbb{R}^{100\times 5023}$ & FLAME shape blendshape parameters, estimated by MLP$_d$ for each Gaussian \\
$\mathcal{W} \in \mathbb{R}^{15\times 5023}$ & FLAME Linear Blend Skinning (LBS) weight, estimated by MLP$_d$ for each Gaussian\\
\midrule
\multicolumn{2}{c}{\bf Rendered and Observed Images}\\
$\hat{\mathbf{I}} / \mathbf{I}$      &   Rendered / Ground Truth Image \\
$\mathbf{M}$      & Mask of subpart region\\
\bottomrule
\end{tabular}
}
\end{table*}

\end{document}